\documentclass[preprint]{article}
\pdfoutput=1

\usepackage{neurips_2024}
\usepackage{natbib}
\setcitestyle{numbers,square}
\usepackage{amsmath}
\usepackage{amssymb}
\usepackage{mathtools}
\usepackage{amsthm}
\usepackage{wrapfig}

\usepackage{float}
\usepackage{caption}

\theoremstyle{plain}
\newtheorem{theorem}{Theorem}[section]

\theoremstyle{definition}

\theoremstyle{remark}

\newcommand{\mr}[2]{\multirow{#1}{*}{#2}}
\newcommand{\mc}[3]{\multicolumn{#1}{#2}{#3}}

\usepackage[utf8]{inputenc} 
\usepackage[T1]{fontenc}    
\usepackage{hyperref}       
\usepackage{url}            
\usepackage{booktabs}       
\usepackage{amsfonts}       
\usepackage{nicefrac}       
\usepackage{microtype}      
\usepackage{xcolor}         
\usepackage{color}
\usepackage{graphicx}
\usepackage{subfigure}
\usepackage{subcaption}
\usepackage{bbding}
\usepackage{booktabs}
\usepackage{amsmath}
\usepackage{amssymb}
\usepackage{mathtools}
\usepackage{amsthm}
\usepackage{enumitem}
\usepackage{multirow}

\title{ROPO: Robust Preference Optimization\\for Large Language Models}

\author{%
    Xize Liang\thanks{Equal Contribution.}~~\thanks{Work done during Xize Liang's research internship at Alibaba Cloud.}~~$^1$ \quad\quad Chao Chen\footnotemark[1]~~$^2$ \quad\quad Shuang Qiu\footnotemark[1]~~$^3$ \quad\quad Jie Wang\thanks{Corresponding Author.}~~$^1$ \quad\quad Yue Wu$^2$ \\
    \textbf{Zhihang Fu$^2$ \quad\quad Zhihao Shi$^1$ \quad\quad Feng Wu$^1$ \quad\quad Jieping Ye$^2$} \\
    $^1$University of Science and Technology of China \quad\quad $^2$Alibaba Cloud \quad\quad $^3$HKUST \\
    \texttt{\{xizeliang, zhihaoshi\}@mail.ustc.edu.cn}\\
    \texttt{masqiu@ust.hk}\quad\quad \texttt{\{jiewangx, fengwu\}@ustc.edu.cn} \\
    \texttt{\{ercong.cc, matthew.wy, zhihang.fzh, yejieping\}@alibaba-inc.com}
}

\begin{document}

\maketitle


\begin{abstract}
    Preference alignment is pivotal for empowering large language models (LLMs) to generate helpful and harmless responses.
    However, the performance of preference alignment is highly sensitive to the prevalent noise in the preference data.
    Recent efforts for this problem either marginally alleviate the impact of noise without the ability to actually reduce its presence, or rely on costly teacher LLMs prone to reward misgeneralization.
    To address these challenges, we propose the \textbf{RO}bust \textbf{P}reference \textbf{O}ptimization (\textbf{ROPO}) framework, an iterative alignment approach that integrates \textit{noise-tolerance} and \textit{filtering of noisy samples} without the aid of external models.
    Specifically, ROPO iteratively solves a constrained optimization problem, where we dynamically assign a quality-aware weight for each sample and constrain the sum of the weights to the number of samples we intend to retain.
    For noise-tolerant training and effective noise identification, we derive a robust loss by suppressing the gradients of samples with high uncertainty.
    We demonstrate both empirically and theoretically that the derived loss is critical for distinguishing noisy samples from clean ones.
    Furthermore, inspired by our derived loss, we propose a robustness-guided rejection sampling technique to compensate for the potential important information in discarded queries.
    Experiments on three widely-used datasets with Mistral-7B and Llama-2-7B demonstrate that ROPO significantly outperforms existing preference alignment methods, with its superiority growing as the noise rate increases.
\end{abstract}

\section{Introduction}

Recent research indicates that the significant achievements of Large Language Models (LLMs) in understanding various queries and providing helpful responses \cite{achiam2023gpt} rely on the preference alignment, which aligns LLMs' responses with human values and expectations \cite{wang2023aligning-v2, bubeck2023sparks, lin2023unlocking}.
A typical preference alignment approach is Reinforcement Learning from Human Feedback (RLHF) \cite{casper2023open, ziegler2019fine}, which first trains a reward model to fit human preferences and subsequently employs an RL algorithm \citep{schulman2017proximal} to guide LLMs to generate high-reward responses.
However, due to the potential risks of misgeneralized reward modeling \citep{casper2023open} and the unstable training \citep{liu2023chain, shen2023large} of RLHF, various ranking-based methods represented by Direct Preference Optimization (DPO) \cite{dpo} bypass the explicit reward modeling stage and eschew RL techniques via directly optimizing the implicit reward margins between preferred and dis-preferred responses \citep{rlhf-complex-2, f-dpo, song2023preference}.
Owing to the stable and computationally lightweight supervised learning paradigm, ranking-based methods have emerged as competitive alternatives to RLHF, thus drawing increasing attention recently \citep{shen2023large, wang2023aligning-v2}.

Despite their impressive performance on preference alignment, ranking-based methods heavily rely on the high-quality preference data, which is costly and limited in practice \citep{kim2023aligning, chen2024self}.
First, the noise (e.g., incorrect or ambiguous preferences) in the preference data is unavoidable \citep{wang2024secrets}.
Many recent studies have observed the presence of preference noise at levels of 20\%-40\% across various scenarios \citep{gao2024impact, lee2023rlaif, zheng2024judging, llama2, cui2023ultrafeedback, zhao2023slic, munos2023nash}, whether the annotators are humans or LLMs.
Second, the performance of LLMs will significantly deteriorate when trained with noisy preferences \citep{rdpo, gao2024impact, lee2023rlaif}.
For instance, a 10\% increase in the noise rate may lead to a 30\% decrease in the performance of DPO in terms of win rate \citep{gao2024impact}.
Therefore, it is highly desirable to develop noise-robust preference alignment techniques.

To address these problems, some recent studies have explored the label smoothing \citep{rdpo, cdpo} and regularization \citep{gao2024impact} techniques to alleviate the impact of preference noise.
However, these methods can only marginally mitigate the side effects of noise, as the noisy data is still involved in the training phase.
Besides, \cite{gao2024impact} also attempts to filter out noisy samples but requires another teacher LLM (i.e., a reward model serving as the proxy of the Bradley-Terry model \cite{bt_model}) to assign confidence values to samples, which introduces additional computational costs.
Moreover, the teacher LLM may not necessarily provide the correct preference direction on some specific domain \citep{casper2023open} and this method is shown to be ineffective at reducing random symmetric noise \citep{gao2024impact}.

In this paper, we propose the \textbf{RO}bust \textbf{P}reference \textbf{O}ptimization (\textbf{ROPO}) framework, an iterative alignment approach that unifies \textit{noise-tolerance} and \textit{filtering of noisy samples} without the aid of external models.
We first provide a general formulation of learning from noisy preference data as a constrained optimization problem, where we dynamically assign a quality-aware weight for each sample.
Then, we solve the problem through an provably convergent iterative paradigm, consisting of two alternating steps: \textit{noise-tolerant model training} and \textit{noisy sample filtering}.
The main contributions of our method are as follows.
\begin{itemize}[leftmargin=10pt]
    \item We propose a robust preference alignment framework that unifies noise-tolerance and filtering of noisy samples.
    Without the need for any external LLM, the model's robustness and discrimination ability against noisy samples gradually improve as the alternating iterative training proceeds.

    \item We derive a robust loss function by suppressing the gradients of samples with high uncertainty.
    The loss contains a regularizer, which not only prevents the model from over-fitting to noisy samples, but also facilitates identifying noisy samples versus clean samples\footnote{In Section \ref{sec:loss}, we demonstrate that the cross-entropy loss (i.e., DPO loss) cannot distinguish between noisy samples and clean samples in the context of preference learning, even though it is widely-used for learning from noisy data in other scenarios such as image classification \cite{mentornet, liu2020early}.} (see Section \ref{sec:loss}).

    \item We propose a robustness-guided rejection sampling technique to compensate for the potential important information in discarded queries (see Section \ref{sec:online}), which improves the data quality and thus leads to further improvement in alignment performance.

    \item We conduct extensive experiments on three widely-used datasets (i.e., UltraFeedback Binarized, Alpaca Comparison, and TL;DR) with Mistral-7B and Llama-2-7B.
    The results show that the performance of ROPO remains stable in noisy scenarios, experiencing a decrease of up to 4\% in the win rate under even 40\% artificial noise (see Table \ref{tab:sft}).
\end{itemize}

\vspace{-1mm}
\section{Preliminaries and Problem Settings}
\vspace{-1mm}
Given a query $\mathbf{x} = [x_1, \ldots, x_n]$, an LLM $\pi_\theta$ (with parameters $\theta$) generates a response $\mathbf{y} = [y_1, \ldots, y_m]$, where the tokens $(x_i)_{i=1}^n$ and $(y_j)_{j=1}^m$ come from a predefined vocabulary, in an autoregressive paradigm.
Specifically, the model samples $y_j$ from the conditional probability distribution $\pi_\theta(\cdot \mid \mathbf{x},\mathbf{y}_{1:j-1})$, where $\mathbf{y}_{1: 0}$ is null and $\mathbf{y}_{1: j-1} = [y_1,\ldots,y_{j-1}]$ for $j=2,\ldots,m$.
Finally, we can decompose the conditional probability $\pi_{\theta}(\mathbf{y} \mid \mathbf{x})$ into $\pi_{\theta}(\mathbf{y} \mid \mathbf{x}) = \prod_{j=1}^m \pi_\theta(y_j \mid \mathbf{x},\mathbf{y}_{1:j-1})$.

\subsection{Alignment of Large Language Models}\label{sec:preference_alignment}

Most of existing LLM alignment frameworks first fine-tune a pre-trained model on high-quality datasets of downstream tasks (e.g., dialogue and post summarization) via maximum likelihood, in order to teach the model to respond to queries.
We denote the supervised fine-tuned model $\pi_{\rm sft}$.
Then, we train the model $\pi_\theta$ (initialized by $\pi_{\rm sft}$) based on human preference data.
Specifically, a preference sample contains a query $\mathbf{x}$, responses $\mathbf{y}_1$ and $\mathbf{y}_2$, and a ranking label $c$ provided annotators.
We use $c=0$ to indicate that $\mathbf{y}_1$ is preferred to $\mathbf{y}_2$ (denoted $\mathbf{y}_1 \succ \mathbf{y}_2 \mid \mathbf{x}$) and use $c=1$ to indicate the opposite.
We assume that the preference data $(\mathbf{x}, \mathbf{y}_1, \mathbf{y_2}, c)$ is sampled from a distribution $\mathcal{D}$.

A popular formulation of the generation of preferences is the Bradley-Terry (BT) model \cite{bt_model}, i.e., $P^*(\mathbf{y}_1 \succ \mathbf{y}_2 \mid \mathbf{x}) = \sigma(r^*(\mathbf{y}_1,\mathbf{x}) - r^*(\mathbf{y}_2,\mathbf{x}))$, where $\sigma$ is the sigmoid function, and $r^*$ is a latent and inaccessible reward function.
The key to existing preference learning methods is to explicitly or implicitly approximate $r^*$ or $P^*$.
RLHF \cite{ouyang2022training} approximates $r^*$ by training a parameterized reward model $r_\phi$ via maximum likelihood on preference data,
then uses the well-trained $r_\phi$ to provide signal for the reinforcement learning of $\pi_\theta$.

Due to the complexity and instability of RLHF, some recent works \cite{dpo, ipo, f-dpo} directly learn human preferences from offline collected response pairs by optimizing the implicit reward margins between preferred and dis-preferred responses.
For example, the objectives of DPO \cite{dpo} is given by
\begin{align}\label{eqn:dpo_loss_pre}
    \ell_{\rm dpo} = -\log \sigma\bigg(\beta \log \frac{\pi_{\theta}(\mathbf{y}_1 \mid \mathbf{x})}{\pi_{\rm ref}(\mathbf{y}_1 \mid \mathbf{x})} - \beta \log \frac{\pi_{\theta}(\mathbf{y}_2 \mid \mathbf{x})}{\pi_{\rm ref}(\mathbf{y}_2 \mid \mathbf{x})}\bigg),
\end{align}
where $\mathbf{y}_1 \succ \mathbf{y}_2 \mid \mathbf{x}$, $\beta$ is a hyperparameter, and $\pi_{\rm ref}$ is a fixed reference model.
Ranking-based methods are more computationally lightweight and stable than RLHF, thus drawing increasing attention recently.
Therefore, we mainly focus on ranking-based methods in this paper.

\begin{figure*}[!t]
    \centering
    \includegraphics[width=\textwidth]{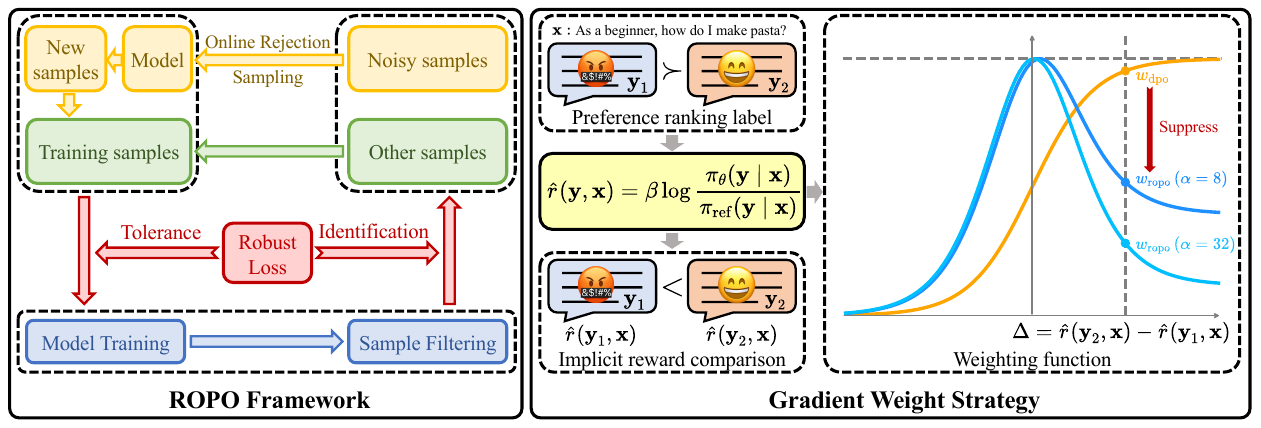}
    \caption{
        Framework of ROPO and a comparison between the gradient weighting strategies of ROPO and DPO \cite{dpo}.
        \textbf{Left:} ROPO alternates between noise-tolerant model training and noisy sample filtering, and integrates the online rejection sampling paradigm to further improve the data quality.
        \textbf{Right:} Unlike $w_{\rm dpo}$, which increases monotonically with respect to $\Delta = \hat{r}(\mathbf{y}_2,\mathbf{x}) - \hat{r}(\mathbf{y}_1, \mathbf{x})$, $w_{\rm ropo}$ decreases when $\Delta$ is large.
        Given a noisy sample $(\mathbf{x}, \mathbf{y}_1, \mathbf{y}_2, \mathbf{y}_1 \succ \mathbf{y}_2 \mid \mathbf{x})$, whose preference ranking label contradicts the comparison of implicit rewards, ROPO suppresses its gradient.
        A larger value of $\alpha$ implies a stronger suppressive effect.
    }
    \label{fig:ropo}
\end{figure*}
\vspace{-2mm}
\subsection{Preference Learning with Noisy Data}\label{sec:pre_noisy}

Preference labels are unavoidably noisy due to the inherent cognitive bias among different annotators.
Therefore, we have no access to the clean dataset, denoted $D = \{(\mathbf{x}^{(i)}, \mathbf{y}_1^{(i)}, \mathbf{y}_2^{(i)}, c^{(i)})\}_{i=1}^N \sim \mathcal{D}$.
Instead, we can only obtain a noisy dataset $D_{\eta} = \{(\mathbf{x}^{(i)}, \mathbf{y}_1^{(i)}, \mathbf{y}_2^{(i)}, \hat{c}^{(i)})\}_{i=1}^N \sim \mathcal{D}_\eta$, where $\hat{c}^{(i)} = c^{(i)}$ with probability $1- \eta$ and $\hat{c}^{(i)} = 1-c^{(i)}$ with probability $\eta$.
Here we assume the random symmetric noise because it is the standard assumption for learning from noisy data \cite{liu2020peer, zhang2018generalized} and existing research has indicated the challenges posed by this kind of noise \cite{gao2024impact}.
Besides, in the context of preference alignment, the symmetric and asymmetric (or class-conditional) noise is equivalent, as the ground truth label is changed if we swap the positions of $\mathbf{y}_1$ and $\mathbf{y}_2$.

\vspace{-2mm}
\section{Robust Preference Optimization}\label{sec:method}
\vspace{-1mm}
We propose \textbf{RO}bust \textbf{P}reference \textbf{O}ptimization (\textbf{ROPO}), an iterative preference alignment framework.
ROPO alternates between \textit{noise-tolerant model training} and \textit{noisy sample filtering}, as shown in Figure \ref{fig:ropo}, which is mathematically equivalent to iteratively solving a constrained optimization problem (Section \ref{sec:formulation}).
In the model training step, we introduce a robust loss function by suppressing the gradients of samples with high uncertainty, which prevents the model from over-fitting to the noisy preference.
In the sample filtering step, we filter out noisy samples based on the magnitude of their training losses.
The key to ROPO is that our proposed loss contains a regularizer, which not only features noise-tolerance, but also facilitates identifying noisy samples versus clean samples (Section \ref{sec:loss}).
Further, we propose a robustness-guided rejection sampling technique to compensate for the potential important information in discarded queries and thus improve the data quality (Section \ref{sec:online}).
\textbf{For detailed proofs of the theorems in this section, please refer to Appendix \ref{append:math}.}

\subsection{A General Formulation}\label{sec:formulation}
Given $N$ preference samples $\{(\mathbf{x}^{(i)}, \mathbf{y}_1^{(i)}, \mathbf{y}_2^{(i)}, \hat{c}^{(i)})\}_{i=1}^N$, we hope that the weights of noisy samples in the preference optimization are smaller than those of others, thereby reducing the impact of noise on the alignment performance.
Without prior knowledge of which samples are noisy, a natural approach would be to assign a dynamic quality-aware weight to each sample and constrain the sum of these weights to a constant, which can also prevent the weights from tending towards zero.
Therefore, we formulate learning from noisy preference samples as the following constrained optimization problem:
\begin{align}
    \min_{\theta, \mathbf{w}} \quad& \frac{1}{N} \sum_{i=1}^N w_i \ell \left(\theta; \mathbf{x}^{(i)}, \mathbf{y}_1^{(i)}, \mathbf{y}_2^{(i)}, \hat{c}^{(i)}, \pi_\theta\right),\label{eqn:primal_problem}\\
    {\rm s.t.} \quad& \theta \in \Theta, \quad w_i \in [0,1],\, i=1,\ldots,N,\quad \sum_{i=1}^N w_i = N_\rho \triangleq \lfloor(1-\rho)N\rfloor,\nonumber
\end{align}
where $w_1, \ldots, w_N$ are dynamic weights, $\Theta$ is compact, and $\rho \in [0,1]$ is the proportion of the samples we aim to filter out.
Please note that we minimize Problem (\ref{eqn:primal_problem}) with respect to both $\theta$ and $\mathbf{w}$, resulting in a training process that learns the weights adaptively.
Hence, we expect that Problem (\ref{eqn:primal_problem}) will gradually lead to much smaller weights for noisy samples than those for others.
To achieve this, we first analyze the properties of the optimal solution to Problem (\ref{eqn:primal_problem}).
As shown in Theorem \ref{thm:primal_problem}, Problem (\ref{eqn:primal_problem}) admits an optimal solution and the elements in its minimizer $\mathbf{w}^*$ are either $0$ or $1$.

\begin{theorem}\label{thm:primal_problem}
    Assume that the loss function $\ell(\theta)$ is continuous on an compact parameter space $\Theta$, then Problem (\ref{eqn:primal_problem}) admits an optimal solution $(\theta^*, \mathbf{w}^*)$.
    Suppose that
    \begin{align}
        \ell\left(\theta^*;\mathbf{x}^{(i_1)}, \mathbf{y}^{(i_1)}_1, \mathbf{y}^{(i_1)}_2, \pi_{\theta^*}\right) < \cdots < \ell\left(\theta^*;\mathbf{x}^{(i_N)}, \mathbf{y}^{(i_N)}_1, \mathbf{y}^{(i_N)}_2, \pi_{\theta^*}\right),
    \end{align}
    then we have $w^*_{i_k} = 1$ for $1\le k \le N_\rho$ and $w^*_{i_k} = 0$ for $N_\rho < k \le N$.
\end{theorem}

We solve Problem (\ref{eqn:primal_problem}) in an iterative paradigm, which consists of two alternating steps: model training and sample filtering.
In the step of model training, we fix $\mathbf{w}$ and learn model parameters $\theta$.
In the step of sample filtering, we fix $\theta$ and assign weights $w_1, \ldots, w_N$ for samples based on their loss values.
Because the objective in Problem (\ref{eqn:primal_problem}) is non-negative and its value does not increase during the iteration, the iterative solving process is guaranteed to converge.

\subsection{A Noise-Tolerant Loss}\label{sec:loss}

To guarantee the effectiveness of the iterative solving process within the preference alignment framework, we delve into identifying additional conditions that should be imposed on $\ell$.
Here, we discuss the properties of $\ell$ in the context of minimizing its expected risks under distributions of noisy and clean preference data, i.e., finding the optimal solutions $\theta^*$ and $\theta^*_\eta$ by solving
\begin{align}\label{eqn:clean_min}
    \theta^* = \mathop{\arg\min}\limits_{\theta \in \Theta} \,\, \mathbb{E}_{(\mathbf{x}, \mathbf{y}_1, \mathbf{y}_2, c)\sim \mathcal{D}}[\ell(\theta;\mathbf{x}, \mathbf{y}_1, \mathbf{y}_2, c,\pi_\theta)]
\end{align}
and
\begin{align}\label{eqn:noisy_min}
    \theta^*_\eta = \mathop{\arg\min}\limits_{\theta \in \Theta} \,\,\mathbb{E}_{(\mathbf{x}, \mathbf{y}_1, \mathbf{y}_2, \hat{c})\sim \mathcal{D}_\eta}[\ell(\theta;\mathbf{x}, \mathbf{y}_1, \mathbf{y}_2, \hat{c},\pi_\theta)],
\end{align}
respectively, where $\Theta$ is a compact parameter space.

\textbf{Requirement 1: Noise-tolerance.}
It cannot be guaranteed that the sample filtering stage will eliminate all noise samples (e.g., when $\rho$ is less than the actual noise proportion in the preference data).
Consequently, it is crucial that the presence of noisy preferences does not significantly impact the model training stage, i.e., $\ell$ is noise-tolerant.

\textbf{Requirement 2: Distinguishable losses for clean and noisy samples.}
As noisy samples generally exhibit larger loss values \cite{liu2020early}, in the sample filtering step, we filter out the $N - N_\rho$ samples with the largest losses.
It is noteworthy that this step takes place midway through training, hence $\ell$ needs to exhibit distinguishable loss values for clean and noisy samples prior to the convergence of the model.

As DPO is one of the most popular preference alignment methods, it is natural for us to explore the effectiveness of the DPO loss $\ell_{\rm dpo}$ (as shown in Eq. (\ref{eqn:dpo_loss_pre})) within our iterative solving process.
However, our findings show that $\ell_{\rm dpo}$ does not satisfy the aforementioned requirements.

\textbf{Finding 1: DPO is not noise-tolerant.}
\begin{theorem}\label{thm:dpo_noise_tolerance}
    Assume $\eta<\frac{1}{2}$.
    Consider $\ell_{\rm dpo}$ and the corresponding minimizer $\theta^*_\eta$ to Problem (\ref{eqn:noisy_min}).
    Given a query $\mathbf{x}$ and responses $(\mathbf{y}_1, \mathbf{y}_2)$, the relationship between the preference probability given by the optimal model, i.e., $P_{\theta^*_\eta}(\mathbf{y}_1 \succ \mathbf{y}_2 \mid \mathbf{x})$, and that given by the BT model, i.e., $P^*(\mathbf{y}_1 \succ \mathbf{y}_2 \mid \mathbf{x})$ is
    \begin{align}
        &P_{\theta^*_\eta}(\mathbf{y}_1 \succ \mathbf{y}_2 \mid \mathbf{x}) = P^*(\mathbf{y}_1 \succ \mathbf{y}_2 \mid \mathbf{x}) + \left( 1 - 2 P^*(\mathbf{y}_1 \succ \mathbf{y}_2 \mid \mathbf{x})\right) \cdot \eta,
    \end{align}
    hence we have
    \begin{align}\label{eqn:dpo_noise_tolerance}
        \big|P_{\theta^*_\eta}(\mathbf{y}_1 \succ \mathbf{y}_2 \mid \mathbf{x}) - P_{\theta^*}(\mathbf{y}_1 \succ \mathbf{y}_2 \mid \mathbf{x})\big| = 2 \eta \big| P^*(\mathbf{y}_1 \succ \mathbf{y}_2 \mid \mathbf{x}) - 1/2\big|.
    \end{align}
\end{theorem}
As shown in Theorem \ref{thm:dpo_noise_tolerance}, the impact of noise on the optimal solution corresponding to $\ell_{\rm dpo}$ increases as the noise rate increases.
Specifically, Eq. (\ref{eqn:dpo_noise_tolerance}) shows that the difference between the optimal probabilities under noisy and clean distributions, i.e., $\big|P_{\theta^*_\eta}(\mathbf{y}_1 \succ \mathbf{y}_2 \mid \mathbf{x}) - P_{\theta^*}(\mathbf{y}_1 \succ \mathbf{y}_2 \mid \mathbf{x})\big|$, is proportional to the label flipping probability $\eta$.

\textbf{Finding 2: DPO faces challenges in distinguishing between noisy and clean samples.}
\begin{theorem}\label{thm:dpo_filter}
    Assume $\eta<\frac{1}{2}$.
    Consider $\ell_{\rm dpo}$ and the corresponding minimizer $\theta^*_\eta$ to Problem (\ref{eqn:noisy_min}).
    For samples $(\mathbf{x}^{(1)}, \mathbf{y}^{(1)}_1, \mathbf{y}^{(1)}_2, \hat{c}^{(1)} = c^{(1)})$ and $(\mathbf{x}^{(2)}, \mathbf{y}^{(2)}_1, \mathbf{y}^{(2)}_2, \hat{c}^{(2)} = 1 - c^{(2)})$, suppose that $\theta$ is not $\theta^*_\eta$ but satisfies $\max\limits_{i=1,2} \bigg|P_\theta\left(\mathbf{y}_1^{(i)} \succ \mathbf{y}_2^{(i)} \mid \mathbf{x}^{(i)}\right) - P_{\theta^*_\eta}\left(\mathbf{y}^{(i)}_1 \succ \mathbf{y}^{(i)}_2 \mid \mathbf{x}^{(i)}\right)\bigg|  < \delta$, then if we want to ensure that $\ell_{\rm dpo}\left(\mathbf{x}^{(1)}, \mathbf{y}^{(1)}_1, \mathbf{y}^{(1)}_2, \hat{c}^{(1)}\right) < \ell_{\rm dpo}\left(\mathbf{x}^{(2)}, \mathbf{y}^{(2)}_1, \mathbf{y}^{(2)}_2, \hat{c}^{(2)}\right)$, $\delta$ must satisfy
    \begin{align}\label{eqn:dpo_filter}
        \delta < \frac{1 - 2\eta}{2}\left( P^*(c^{(1)}) + P^*(c^{(2)}) - 1 \right).
    \end{align}
\end{theorem}
As shown in Theorem \ref{thm:dpo_filter}, the distance between $\pi_{\theta}$ and $\pi_{\theta^*_\eta}$ we need for $\ell_{\rm dpo}$ to differentiate between clean and noisy samples decreases as the BT probability approaches 50\% and the noise rate increases.
Specifically, Eq. (\ref{eqn:dpo_filter}) shows that the upper bound of $\delta$ is proportional to $(1-2\eta)/2$ and $\left( P^*(c^{(1)}) - 1/2 + P^*(c^{(2)}) - 1/2 \right)$\footnote{As $c^{(1)}$ and $c^{(2)}$ are clean labels, we have $P^*(c^{(1)})>1/2$ and $P^*(c^{(2)}) > 1/2$.}.
Due to the intrinsic diversity and stochastic nature of human preferences, the BT distribution is usually not a ``hard'' distribution with probabilities close to 0 or 1, but rather a ``soft'' one \cite{spo, strobl2011accounting}.
This brings difficulties to unconverged model trained with DPO in distinguishing between noisy and clean samples.
For example, when $\eta=30\%$ and $P^*\left(c^{(1)}\right) = P^*\left(c^{(2)}\right) = 60\%$, we need $\delta < 4\%$, which is a challenging requirement for a model that has not yet converged.

\textbf{The gradient weighting strategy of DPO may amplify the impact of noise.}
Given a sample $(\mathbf{x}, \mathbf{y}_1, \mathbf{y}_2, \hat{c}=0)$, according to \cite{dpo}, the gradient of $\ell_{\rm dpo}$ in Eq. (\ref{eqn:dpo_loss_pre}) is given by
\begin{align}\label{eqn:dpo_gradient}
    \nabla_\theta \ell_{\rm dpo} = - \beta \underbrace{ \sigma\left( \hat{r}(\mathbf{y}_2, \mathbf{x}) - \hat{r}(\mathbf{y}_1, \mathbf{x}) \right)}_{w_{\rm dpo}(\mathbf{x}, \mathbf{y}_1, \mathbf{y}_2)} \cdot \nabla \log \frac{\pi_\theta(\mathbf{y}_1\mid \mathbf{x})}{\pi_\theta(\mathbf{y}_2 \mid \mathbf{x})},
\end{align}
where $\hat{r}(\mathbf{y}, \mathbf{x}) = \beta \log \frac{\pi_\theta (\mathbf{y} \mid \mathbf{x})}{\pi_{\rm ref}(\mathbf{y} \mid \mathbf{x})}$ is the implicit reward function of DPO.
Intuitively, the greater the discrepancy between the reward function's comparison of $\mathbf{y}_1$ and $\mathbf{y}_2$ and the label $\mathbf{y}_1 \succ \mathbf{y}_2 \mid \mathbf{x}$, the greater the weight $w_{\rm dpo}(\mathbf{x}, \mathbf{y}_1, \mathbf{y}_2)$ of the DPO gradient becomes.
This aggressive weighting strategy can be risky if the label is incorrect, as the model may imply a high uncertainty about the sample by giving a higher reward to $\mathbf{y}_2$ than to $\mathbf{y}_1$, increasing $w_{\rm dpo}$ and thus amplifying the impact of the noise.

\textbf{Conservative gradient weighting strategy.}
A simple and straightforward idea is that when the implicit reward margin $\Delta(\mathbf{y}_2, \mathbf{y_1}, \mathbf{x})\triangleq\hat{r}(\mathbf{y}_2, \mathbf{x}) - \hat{r}(\mathbf{y}_1, \mathbf{x})$ is excessively positive, we should assign a conservative weight to the gradient.
Based on this idea, we propose the conservative gradient weight
\begin{align}
    w_{\rm ropo} = \frac{4\alpha}{(1+\alpha)^2}\cdot \sigma(\Delta(\mathbf{y}_2, \mathbf{y_1}, \mathbf{x})) \cdot (1 + \alpha \sigma(-\Delta(\mathbf{y}_2, \mathbf{y}_1, \mathbf{x}))),
\end{align}
where $\alpha \ge 1$ controls the conservatism of weighting and $4\alpha / (1+\alpha)^2$ is used to normalize the maximum value of $w_{\rm ropo}$.
As illustrated in Figure \ref{fig:ropo}, unlike the monotonous increase of $w_{\rm dpo}$, $w_{\rm ropo}$ decreases when $\Delta(\mathbf{y}_2, \mathbf{y_1}, \mathbf{x})$ is large.
Then, the corresponding loss function can be decomposed as
\begin{align}\label{eqn:decomposition}
    \ell_{\rm ropo} ={}& \int \nabla_\theta \ell_{\rm ropo}\,{\rm d}\theta = \frac{4\alpha}{(1+\alpha)^2} \cdot \ell_{\rm dpo} + \frac{4\alpha^2}{(1+\alpha)^2} \cdot \ell_{\rm reg},
\end{align}
where $\ell_{\rm reg} = \sigma\left(\beta \log \frac{\pi_\theta (\mathbf{y}_2 \mid \mathbf{x})}{\pi_{\rm ref}(\mathbf{y}_2 \mid \mathbf{x})} - \beta \log \frac{\pi_\theta (\mathbf{y}_1 \mid \mathbf{x})}{\pi_{\rm ref}(\mathbf{y}_1 \mid \mathbf{x})}\right)$ and we omit the constant term of the primitive function (see Appendix \ref{append:decomposition} for the detailed derivation).
The introduced loss consists of $\ell_{\rm dpo}$ and a regularizer $\ell_{\rm reg}$ whose weight is $\alpha$ times that of $\ell_{\rm dpo}$.
We claim that $\ell_{\rm reg}$ has the following advantages.

\begin{figure}[t!]
    \centering
    \subfigure[DPO (ep1). $\gamma=0.48$.]{
        \label{fig:tldr-dpo-no02-llama2-ep1}
        \includegraphics[width=0.233\linewidth]{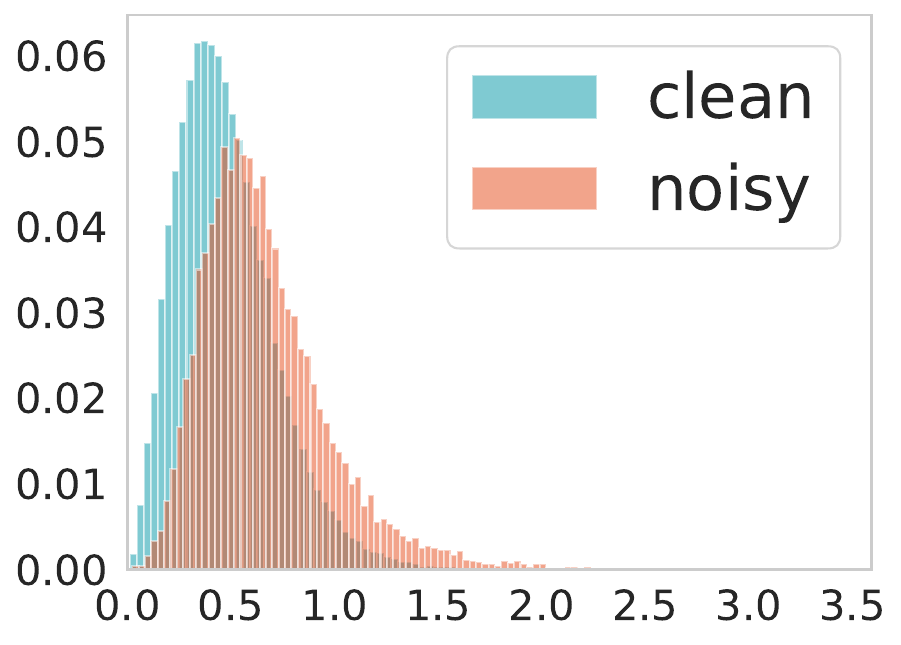}
    }
    \subfigure[DPO (ep2). $\gamma=0.34$.]{
        \label{fig:tldr-dpo-no02-llama2-ep2}
        \includegraphics[width=0.233\linewidth]{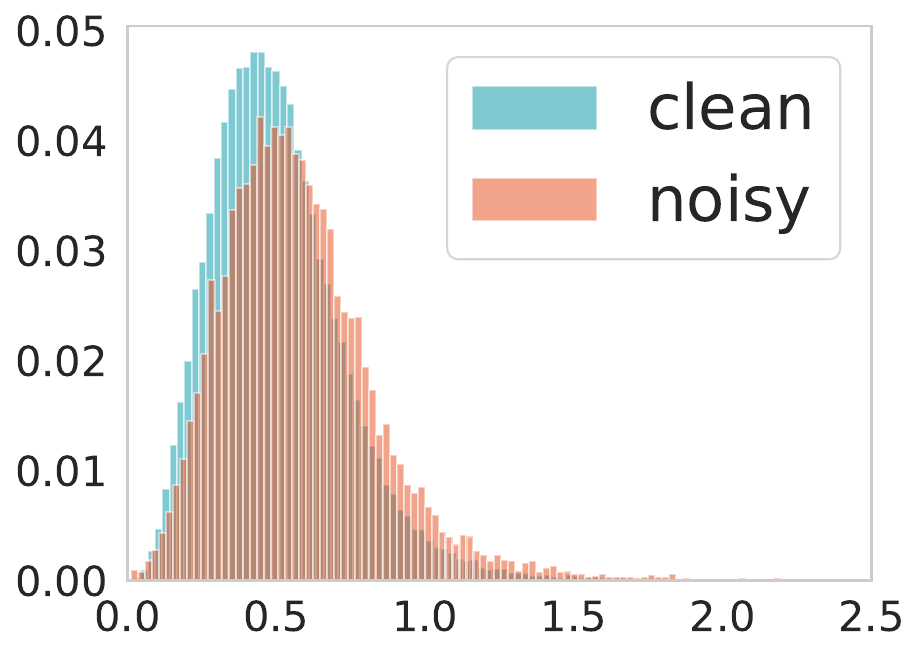}
    }
    \subfigure[ROPO (ep1). $\gamma=0.58$.]{
        \label{fig:tldr-ropo-no02-llama2-ep1}
        \includegraphics[width=0.233\linewidth]{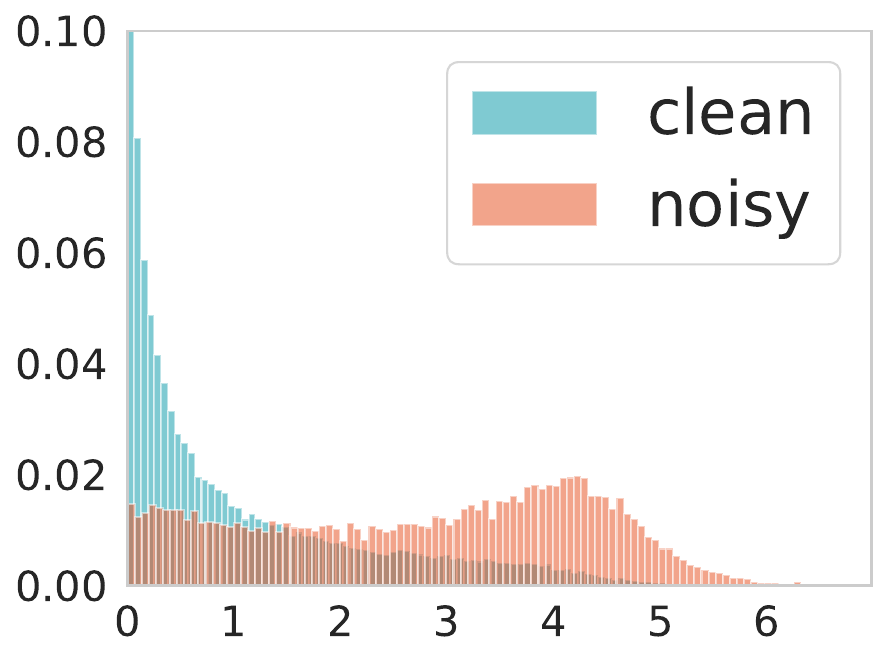}
    }
    \subfigure[ROPO (ep2). $\gamma=0.60$.]{
        \label{fig:tldr-ropo-no02-llama2-ep2}
        \includegraphics[width=0.233\linewidth]{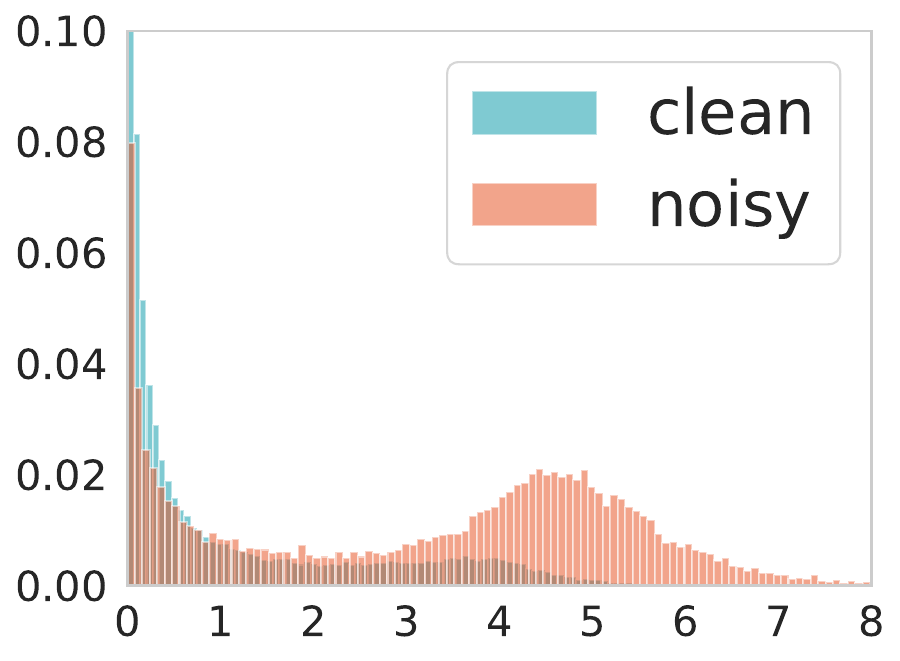}
    }
    \caption{
        Loss distributions of Llama-2-7B trained with DPO and ROPO at different training epochs (ep1 and ep2) on TL;DR.
        We denote $\gamma$ as the proportion of noisy samples in the 20\% of samples that are filtered out.
        Larger $\gamma$ indicates better discrimination between clean and noisy samples.
    }
    \label{fig:loss-distribution}
\end{figure}

\textbf{Advantage 1: $\ell_{\rm reg}$ is noise-tolerant.}
\begin{theorem}\label{thm:reg-robust}
    Assume that $\eta<\frac{1}{2}$.
    Consider $\ell_{\rm reg}$ and the corresponding minimizer $\theta^*_{\eta}$ to Problem (\ref{eqn:noisy_min}).
    Given a query $\mathbf{x}$ and responses $(\mathbf{y}_1, \mathbf{y}_2)$, the relationship between the preference probability given by the optimal model, i.e., $P_{\theta^*_\eta}(\mathbf{y}_1 \succ \mathbf{y}_2 \mid \mathbf{x})$, and that given by the BT model, i.e., $P^*(\mathbf{y}_1 \succ \mathbf{y}_2\mid \mathbf{x})$ is
    \begin{align}\label{eqn:reg-robust}
        P_{\theta^*_\eta}(\mathbf{y}_1 \succ \mathbf{y}_2 \mid \mathbf{x}) = \mathbb{I}\left( P^*(\mathbf{y}_1 \succ \mathbf{y}_2 \mid \mathbf{x}) > \frac{1}{2} \right),
    \end{align}
    hence we have $P_{\theta^*_\eta}(\mathbf{y}_1 \succ \mathbf{y}_2 \mid \mathbf{x}) = P_{\theta^*}(\mathbf{y}_1 \succ \mathbf{y}_2 \mid \mathbf{x})$.
\end{theorem}
As shown in Theorem \ref{thm:reg-robust}, contrary to the conclusion in Theorem \ref{thm:dpo_noise_tolerance} that the optimal solution corresponding to $\ell_{\rm dpo}$ is affected by the noise, the optimal preference probability corresponding to $\ell_{\rm reg}$, i.e., $P_{\theta^*_\eta}(\mathbf{y}_1 \succ \mathbf{y}_2 \mid \mathbf{x})$, remains unchanged when the label flipping probability $\eta<1/2$.
Specifically, Eq. (\ref{eqn:reg-robust}) shows that $P_{\theta^*_\eta}(\mathbf{y}_1 \succ \mathbf{y}_2 \mid \mathbf{x})$ is an indicator function of $P^*(\mathbf{y}_1 \succ \mathbf{y}_2 \mid \mathbf{x})>1/2$.

\textbf{Advantage 2: $\ell_{\rm reg}$ can distinguish noisy samples from clean ones.}
\begin{theorem}\label{thm:reg-dist}
    Assume that $\eta<\frac{1}{2}$.
    Consider $\ell_{\rm reg}$ and the corresponding minimizer $\theta^*_{\eta}$ to Problem (\ref{eqn:noisy_min}).
    For samples $(\mathbf{x}^{(1)}, \mathbf{y}^{(1)}_1, \mathbf{y}^{(1)}_2, \hat{c}^{(1)} = c^{(1)})$ and $(\mathbf{x}^{(2)}, \mathbf{y}^{(2)}_1, \mathbf{y}^{(2)}_2, \hat{c}^{(2)} = 1 - c^{(2)})$, suppose that $\theta$ is not $\theta^*_\eta$ but satisfies $\max\limits_{i=1,2} \bigg|P_\theta\left(\mathbf{y}_1^{(i)} \succ \mathbf{y}_2^{(i)} \mid \mathbf{x}^{(i)}\right) - P_{\theta^*_\eta}\left(\mathbf{y}^{(i)}_1 \succ \mathbf{y}^{(i)}_2 \mid \mathbf{x}^{(i)}\right)\bigg|  < \delta$, then if we want to ensure that $\ell_{\rm reg}\left(\mathbf{x}^{(1)}, \mathbf{y}^{(1)}_1, \mathbf{y}^{(1)}_2, \hat{c}^{(1)}\right) < \ell_{\rm reg}\left(\mathbf{x}^{(2)}, \mathbf{y}^{(2)}_1, \mathbf{y}^{(2)}_2, \hat{c}^{(2)}\right)$, we must have $\delta < \frac{1}{2}$.
\end{theorem}
As shown in Theorem \ref{thm:reg-dist}, contrary to the challenging requirement $\ell_{\rm dpo}$ places on an unconverged model in Theorem \ref{thm:dpo_filter}, we can expect that $\ell_{\rm reg}$ yields a larger value for noisy samples than for others as long as the difference between the preference probability given by an unconverged model and that of the optimal model is less than 50\%.
We verify our theoretical analysis in experiments, as shown in Figure \ref{fig:loss-distribution}.
For details about the experiments, please refer to Section \ref{sec:exp_main}.

\textbf{Discussion.}
The regularizer $\ell_{\rm reg}$ is capable of improving noise-tolerance and separating noisy samples from clean samples.
Given that the weight of $\ell_{\rm reg}$ is $\alpha \ge 1$ times greater than that of $\ell_{\rm dpo}$, $\ell_{\rm reg}$ dominates the optimization process.
However, $\ell_{\rm dpo}$ leads to a ``softer'' optimal preference probability compared with $\ell_{\rm reg}$, which could potentially avoid discrimination against minorities by LLMs.
Besides, the aggressive weighting strategy may be useful for clean preference datasets (although they are rare).
Thus, it is considered necessary to incorporate a minor component of $\ell_{\rm dpo}$ into the final loss function.
From this perspective, the hyperparameter $\alpha$ plays an important role in trading-off between aggressive ($\ell_{\rm dpo}$) and conservative ($\ell_{\rm reg}$) gradient weighting strategy.

\subsection{Robustness-guided Rejection Sampling}\label{sec:online}

The sample filtering step effectively reduces the proportion of noise but may also discard some important queries.
For example, a query designed to eliminate the occupational discrimination in LLMs may be filtered out because the ranking label of its associated responses is incorrect.
Thus, inspired by the sample distinguishing ability of our proposed $\ell_{\rm ropo}$, we propose a rejection sampling technique to compensate for the essential but discarded information and thus improve the robustness of our ROPO framework.
Specifically, we sample $K$ responses $\widetilde{\mathbf{y}}_{1}, \ldots, \widetilde{\mathbf{y}}_{K}$ to $\mathbf{x}$ for each sample $(\mathbf{x}, \mathbf{y}_1, \mathbf{y}_2)$ that is filtered out and generate $2K$ candidate samples
\begin{align*}
    \{(\mathbf{x}, \mathbf{y}_1, \widetilde{\mathbf{y}}_k, \mathbf{y}_1 \succ \widetilde{\mathbf{y}}_k \mid \mathbf{x})\}_{k=1}^K \cup \{(\mathbf{x}, \mathbf{y}_2, \widetilde{\mathbf{y}}_k, \mathbf{y}_2 \succ \widetilde{\mathbf{y}}_k \mid \mathbf{x})\}_{k=1}^K.
\end{align*}
Then, we compute their loss values and add the sample with the minimum loss to the dataset.
Note that we treat the model's responses as dis-preferred ones compared to the original responses, which on the one hand suppresses the harmful information in its responses, and on the other hand improves the diversity of its outputs.

The rejection sampling is a popular approach of data augmentation to improve the data quality and performance in existing preference alignment methods \cite{dong2023raft, liu2023statistical, xiong2023iterative, yang2024rewards, wang2024arithmetic}.
Specifically, \cite{dong2023raft} ranks newly-collected responses based on their rewards and select the highest ranked one to add to the dataset.
To address the issue of the excessively high rejection rate and thus improve the effectiveness of rejection sampling, \cite{xiong2023iterative} proposes a multi-step sampling technique, which also requires an external reward model.
Besides, \cite{wang2024arithmetic} and \cite{yang2024rewards} consider rejection sampling for the multi-objective preference alignment, where \cite{wang2024arithmetic} projects multi-objective reward vectors onto one dimension and then selects samples based on the scalar rewards, while \cite{yang2024rewards} augments samples near the Pareto front of multi-dimensional rewards, leading to a strong multi-objective alignment performance.
Compared to the aforementioned methods, which all rely on rewards provided by external models, our robustness-guided rejection sampling technique selects new samples based on loss values that reflect the quality of the samples.
Moreover, our technique benefits from being independent of external LLMs, thus leading to computational and memory efficiency.

\section{Experiments}\label{sec:experiments}

\subsection{Experimental Settings}\label{sec:exp_settings}

\textbf{Tasks and Datasets.}
We focus on two dialogue datasets (i.e., UltraFeedback Binarized\footnote{\url{https://huggingface.co/datasets/HuggingFaceH4/ultrafeedback_binarized}} (UFB) and Alpaca Comparison \cite{peng2023instruction}) and one post summarization dataset (i.e., Reddit TL;DR \cite{tldr-1, tldr-2}).
To introduce random noise at different levels, we randomly alter preference labels at different proportions within the three datasets (see Section \ref{sec:exp_main}).
For details about the datasets, please refer to \ref{append:dataset}.

\textbf{Baselines and Models.}
Our baselines are DPO \cite{dpo}, IPO \cite{ipo}, and two approaches using the label smoothing technique to alleviate the impact of noise, i.e., rDPO \cite{rdpo} and cDPO \cite{cdpo}.
We use Mistral-7B \cite{jiang2023mistral} and Llama-2-7B \cite{llama2} as base models for all baselines and datasets.
On UFB, we use Zephyr-7B-SFT-$\beta$ \cite{tunstall2023zephyr} as the SFT model for experiments with Mistral-7B, and adopt the result of Zephyr-7B-$\beta$ \cite{tunstall2023zephyr} on AlpacaEval (90.60) as the performance of DPO under no artificial noise.
In other cases, we fine-tune base models on the preferred responses (SFT targets) to form the SFT models.
For details about our baselines, models, and hyperparamters, please refer to Appendix \ref{append:baselines}.
We run all experiments on 16 NVIDIA A100 GPUs (80 GB).

\textbf{Evaluation.}
For models trained on UFB and Alpaca Comparison, we evaluate them on the AlpacaEval benchmark \cite{alpaca_eval} by comparing their outputs with those of text-davinci-003 (recommended by the benchmark for comparison).
For models trained on TL;DR, we evaluate them by comparing their outputs with the SFT targets (chosen responses) on the test split of TL;DR.
Following existing studies \cite{dpo, tunstall2023zephyr}, we employ GPT-4 as the referee to conduct head-to-head comparisons, using the win rate as the metric.
The win rate can be computed by $\Omega = \frac{\#({\rm Win}) + \#({\rm Tie})/2}{\#({\rm Comparisons})}$, where $\#({\rm Win})$, $\#({\rm Tie})$, and $\#({\rm Comparisons})$ are the numbers of wins, ties, and comparisons, respectively.
For more information about the evaluation, please refer to Appendix \ref{append:evaluation}.

\subsection{Main Results}\label{sec:exp_main}

In this section, we evaluate the performance of ROPO on the three datasets.
The artificial noise is introduced by randomly alter the labels of 20\% or 40\% of samples in the datasets.

\textbf{ROPO is robust to noisy preferences.}
We present the win rates of different methods vs SFT targets under different proportions of artificial noise in Table \ref{tab:sft}.
From the table we can make several interesting observations:
(1) For all preference alignment methods, their win rates show a decreasing trend as the noise rate increases.
(2) Compared to the competitors, our proposed ROPO demonstrates more stable performance under noisy preference data.
Specifically, the win rate of ROPO drops by 2.4\% (at most) under 20\% noisy labels, while those of DPO and IPO drop by 14.4\% and 5.0\%, respectively.
(3) ROPO consistently outperforms the baselines under different proportions of artificial noise in all the three datasets.
Even without artificial noise, ROPO still exceeds DPO by 16.0\% on TL;DR and 5.6\% on Alpaca Comparison, which indicates that the preference dataset inherently includes a significant amount of noise.
(4) Baselines that use the label smoothing technique (i.e., rDPO and cDPO) cannot stably mitigate the impact of noise for DPO.
We speculate that the reasons for their limited effectiveness are as follows.
First, rDPO and cDPO are noise-tolerant only when the hyperparameter $\varepsilon$ exactly equals the proportion of noise and when $\varepsilon=0.5$, respectively (see Appendix \ref{append:rdpo-tolerant}), which is difficult to achieve in practice, as we have no access to the prior knowledge of the exact noise proportion.
Second, they do not reduce the presence of noise, thus can only marginally mitigate the side effects of noise.
In contrast, our ROPO framework exhibits noise-tolerance without requiring the priors on the noise proportion and iteratively reduce the noise proportion as the training proceeds, thus leading to superior performance to rDPO and cDPO.

\begin{table}[!t]
    \centering
    \caption{
        Win rates (\%) of \textbf{different methods vs SFT targets} under different proportions (i.e., 0, 20\%, and 40\%) of artificial noise, evaluated by GPT-4.
        The bold font indicates the best result and an underline indicates the second-best result.
    }
    \label{tab:sft}
    \resizebox{\linewidth}{!}{
        \begin{tabular}{ll|ccc|ccc|ccc}
            \toprule
            \mc{2}{c}{Dataset} & \mc{3}{c}{UFB} & \mc{3}{c}{Alpaca Comparison} & \mc{3}{c}{TL;DR} \\
            \midrule
            Model & Method & 0\% & 20\% & 40\% & 0\% & 20\% & 40\% & 0\% & 20\% & 40\% \\
            \midrule
            \mr{5}{Mistral-7B} & DPO & \underline{90.60} & 86.21 & 82.67 & \underline{69.81} & \underline{68.32} & 65.84 & \underline{63.00} & 56.80 & 49.60 \\
            & IPO & 88.45 & \underline{87.32} & 82.86 & 68.70 & 64.60 & 61.74 & 62.00 & \underline{57.00} & 45.20 \\
            & rDPO & 86.96 & 84.47 & \underline{83.23} & 68.45 & 67.83 & \underline{66.34} & 62.40 & 54.40 & \underline{52.60} \\
            & cDPO & 76.27 & 63.60 & 58.88 & 65.09 & 61.61 & 62.11 & 59.40 & 56.60 & 49.40  \\
            & \textbf{ROPO} & \textbf{91.06} & \textbf{88.63} & \textbf{87.70} & \textbf{75.40} & \textbf{76.27} & \textbf{74.04} & \textbf{79.00} & \textbf{77.80} & \textbf{75.80} \\
            \midrule
            \mr{5}{Llama-2-7B} & DPO & \underline{60.50} & 56.15 & 54.04 & 51.06 & 46.09 & 44.72 & \underline{56.80} & 42.40 & 35.20 \\
            & IPO & 59.75 & 55.40 & 50.06 & 51.80 & 50.56 & 49.57 & 54.20 & \underline{50.80} & \underline{51.60} \\
            & rDPO & 57.76 & \underline{57.02} & \underline{55.90} & 49.81 & 50.19 & \underline{49.94} & 54.80 & 49.00 & 46.00 \\
            & cDPO & 54.04 & 55.40 & 54.16 & \underline{52.17} & \underline{50.81} & 48.20 & 52.20 & 47.40 & 43.00 \\
            & \textbf{ROPO} & \textbf{68.94} & \textbf{69.44} & \textbf{66.71} & \textbf{55.90} & \textbf{54.41} & \textbf{54.53} & \textbf{78.80} & \textbf{78.00} & \textbf{79.20}  \\
            \bottomrule
        \end{tabular}
    }
\end{table}

\begin{table}[!t]
    \centering
    \caption{
        Win rates (\%) of \textbf{ROPO and DPO vs SFT targets} under different proportions (i.e., 0, 20\%, and 40\%) of artificial noise at different training epochs on TL;DR, evaluated by GPT-4.
    }
    \label{tab:grad}
    \resizebox{\linewidth}{!}{
        \begin{tabular}{ll|ccc|ccc|ccc}
            \toprule
            \mr{2}{Model} & \mr{2}{Method} & \mc{3}{c}{0\%} & \mc{3}{c}{20\%} & \mc{3}{c}{40\%} \\
            & & ep1 & ep2 & ep3 & ep1 & ep2 & ep3 & ep1 & ep2 & ep3 \\
            \midrule
            \mr{2}{Mistral-7B} & DPO & 62.60 & 60.20 & 63.00 & 56.80 & 51.00 & 48.60 & 49.60 & 44.40 & 44.60 \\
            & ROPO & 75.40 & 75.60 & 79.00 & 68.80 & 76.40 & 77.80 & 61.60 & 70.80 & 75.80 \\
            \midrule
            \mr{2}{Llama-2-7B} & DPO & 49.00 & 53.60 & 56.80 & 42.40 & 38.40 & 39.20 & 32.00 & 35.20 & 33.60 \\
            & ROPO & 74.00 & 82.00 & 78.80 & 58.40 & 76.40 & 78.00 & 46.00 & 70.80 & 79.20 \\
            \bottomrule
        \end{tabular}
    }
\end{table}

\textbf{ROPO is able to distinguish between noisy and clean samples.}
In Section \ref{sec:loss}, we have theoretically shown that $\ell_{\rm reg}$ can distinguish noisy samples from clean ones, while $\ell_{\rm dpo}$ cannot.
To empirically support our theoretical analysis, we report the distributions of losses for Llama-2-7B trained with ROPO and DPO on TL;DR in Figure \ref{fig:loss-distribution}.
Specifically, for models trained for one (two) epoch, we use the SFT model (the model trained for one epoch) as the reference model and compute the losses for all noisy and clean samples in the training set.
The results in Figure \ref{fig:loss-distribution} demonstrate three important observations:
(1) Our $\ell_{\rm ropo}$ can distinguish between noisy and clean samples by yielding larger values for noisy samples than for others.
(2) The distributions of $\ell_{\rm dpo}$ on noisy and clean samples are similar and the gap between them narrows as training progresses.
(3) ROPO has a stronger capability for filtering out noisy samples compared to DPO.
Specifically, in the top 20\% of samples with the largest $\ell_{\rm ropo}$, noisy samples make up 60\%; whereas in the top 20\% of samples with the largest $\ell_{\rm dpo}$, noisy samples account for about 34\%.

\textbf{ROPO gradually improves the performance.}
In Table \ref{tab:grad}, we report the win rates of ROPO and DPO vs SFT targets under different proportions of artificial noise at different training epochs on TL;DR.
From the results we find that the performance of ROPO gradually improves as training progresses in most cases, while DPO does not exhibit the same trend.
Specifically, the performance of DPO at the second and third epochs is generally lower than that at the first epoch under 20\% and 40\% artificial noise.
As a comparison, the second epoch training of ROPO brings an 8\%-24\% increase in the win rate, and the third epoch also leads to a 5\%-9\% improvement under 40\% artificial noise.
These results demonstrate that the iterative training of ROPO effectively reduces the impact of noise and thus consistently improves the alignment performance.
\begin{wraptable}{!rt}{7cm}
    \caption{
        Ablations on different components of ROPO for Mistral-7B on UFB.
        NSF and RS stand for the noisy sample filtering and rejection sampling stages, respectively.
    }
    \label{tab:ablation}
    \resizebox{\linewidth}{!}{
        \begin{tabular}{l|ccc}
            \toprule
            Method & 0\% & 20\% & 40\% \\
            \midrule
            DPO & 90.60 & 86.21 & 82.67 \\
            ROPO ($\ell_{\rm reg}$) & 89.19 & 87.58 & 86.34 \\
            ROPO ($\ell_{\rm reg}$ + NSF) & 89.44 & 88.20 & \textbf{88.07} \\
            ROPO ($\ell_{\rm reg}$ + NSF + RS) & \textbf{91.06} & \textbf{88.63} & 87.70 \\
            \bottomrule
        \end{tabular}
    }
\end{wraptable}

\subsection{Ablations}\label{sec:ablation}

\textbf{Effectiveness of components in ROPO.}
To evaluate the effectiveness of different components of our ROPO framework, we compare the performance of our proposal with and without: (a) noise-tolerant regularizer $\ell_{\rm reg}$, (b) noisy sample filtering stage, and (c) online rejection sampling stage.
The results are presented in Table \ref{tab:ablation}.
As can be seen, all components improve ROPO's performance, validating the rationale of our robust alignment framework.
Compared to the aggressive DPO loss, the introduced regularizer $\ell_{\rm reg}$ consistently improves the performance, which indicates that a proper trade-off between aggressive and conservative gradient weighting strategy effectively prevents the model from  over-fitting to noise.
Besides, the results also show that the noisy sample filtering is the most effective part of our proposal, which also makes our method significantly superior to other label smoothing-based techniques \cite{rdpo,cdpo}.

\begin{figure}[t!]
    \centering
    \subfigure[Mistral-7B]{
        \label{fig:ablation-rho-mistral}
        \includegraphics[width=0.233\linewidth]{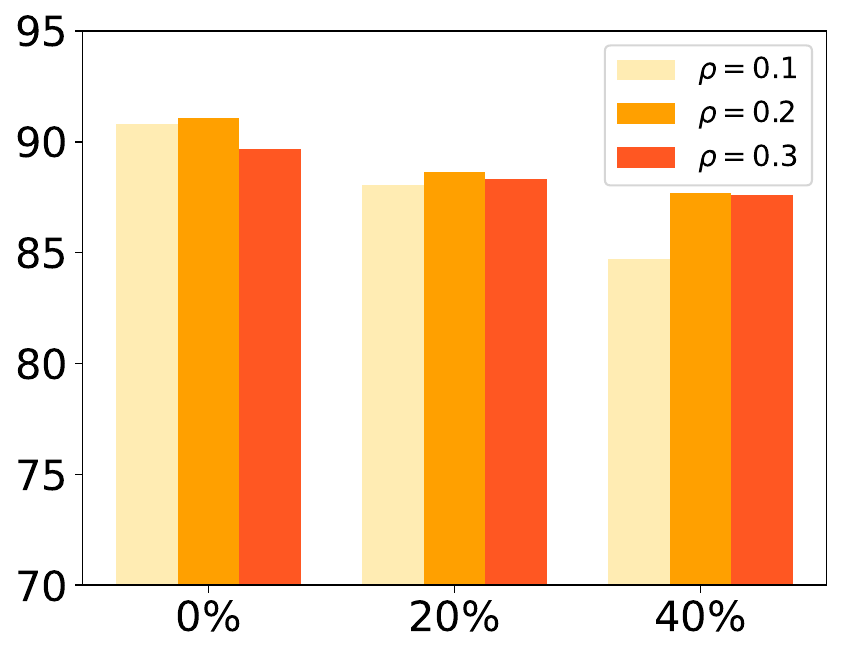}
    }
    \subfigure[Llama-2-7B]{
        \label{fig:ablation-rho-llama2}
        \includegraphics[width=0.233\linewidth]{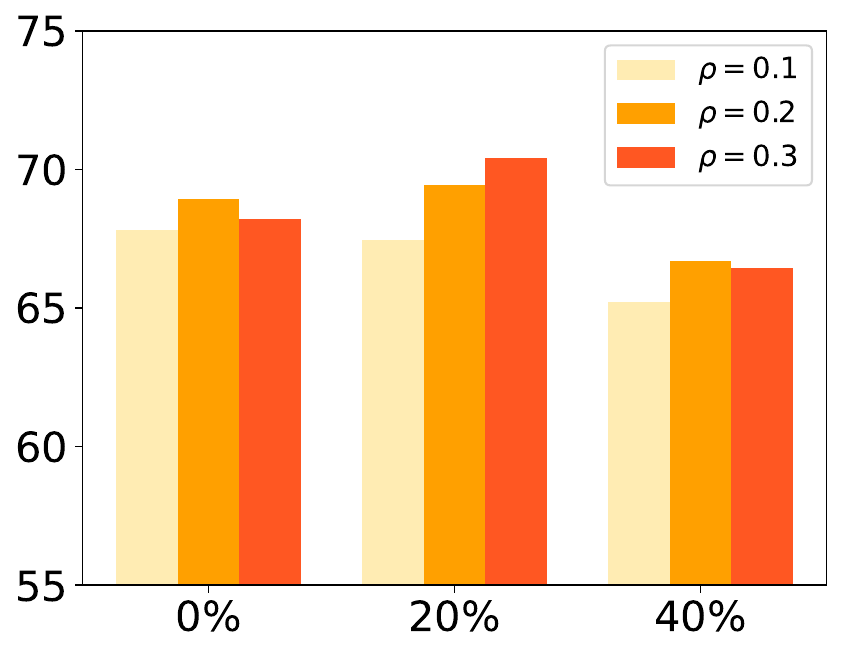}
    }
    \subfigure[UFB]{
        \label{fig:ablation-alpha-ufb}
        \includegraphics[width=0.233\linewidth]{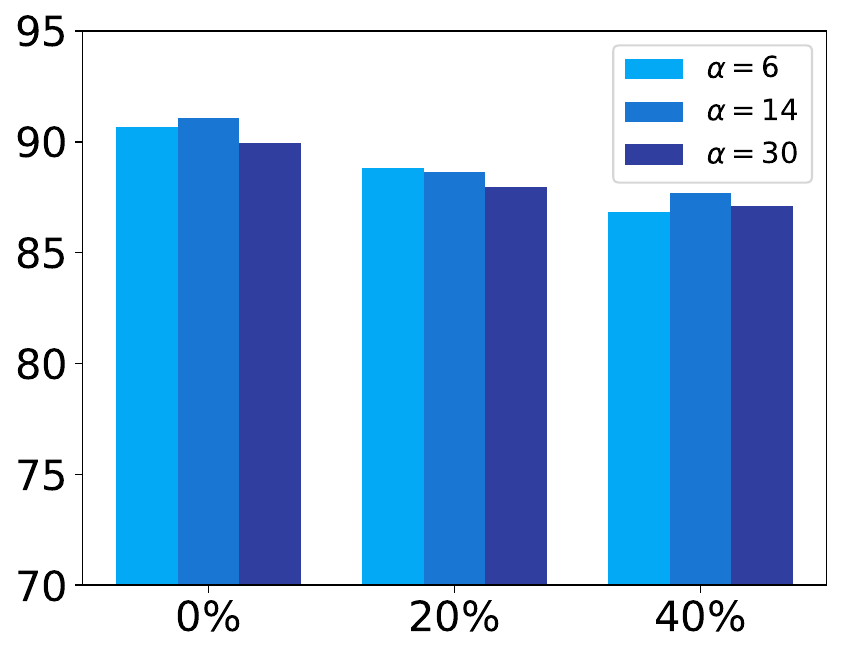}
    }
    \subfigure[TL;DR]{
        \label{fig:ablation-alpha-tldr}
        \includegraphics[width=0.233\linewidth]{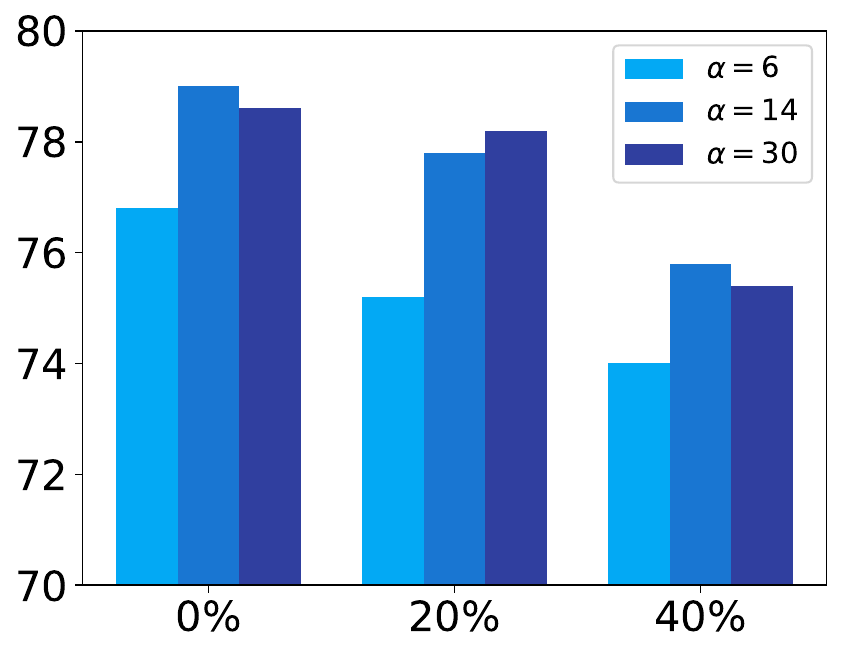}
    }
    \caption{
        Ablations on $\rho$ and $\alpha$.
        (a) and (b) respectively show the performance of ROPO-trained Mistral-7B and Llama-2-7B on UFB with different proportions of artificial noise and sample filtering ratio $\rho$.
        (c) and (d) respectively show the performance of ROPO-trained Mistral-7B on UFB and TL;DR with different proportions of artificial noise and $\alpha$.
    }
    \label{fig:ablation}
\end{figure}

\textbf{How many noisy samples should we filter out?}
The sample filtering ratio $\rho$ is a key factor to the data filtering stage. In the main experiments, we only report the results of filtering 20\% samples. Here, we also present the results of filtering $10\%$ and $30\%$ samples with larger loss values. The results in Fig. \ref{fig:ablation-rho-mistral}, \ref{fig:ablation-rho-llama2} show that better performance could be achieved when filtering 20\% or 30\% samples. We attribute the reason for this result to the noise ratio in the preference data, which is generally between 20\%-30\% \cite{gao2024impact}. There's a substantial risk of eliminating a considerable amount of high-quality data if we set a larger ratio $\rho$.
Thus, we recommend setting the ratio to 20\% in practice.

\textbf{Sensitivity to hyperparameters $\alpha$.}
The trade-off hyperparameter $\alpha$ controls the importance of the conservative regularization term.
A larger $\alpha$ indicates more conservative gradient weighting strategy.
As $C\triangleq \lim_{\Delta \to \infty}w_{\rm ropo}(\Delta) = 4\alpha/(\alpha + 1)^2$, we search the best $\alpha$ in the range of $\{6, 14, 30\}$---which corresponds to $C \in \{1/2, 1/4, 1/8\}$---and use $\alpha=14$ in our main experiments (see Appendix \ref{append:baselines} for the settings of hyperparameters).
To explore the effect of $\alpha$, we provide ablations on $\alpha$ in Figures \ref{fig:ablation-alpha-tldr} and \ref{fig:ablation-alpha-ufb}.
As observed, the model's performance remains largely unaffected for $\alpha$ within an appropriate range, as the loss scale does not change significantly (note that $\alpha C \in [2.94, 3.75]$ for $\alpha \in [6, 30]$).
Besides, for the dialogue task, a smaller $\alpha$ results in better performance, as a smaller $\alpha$ will lead to more diverse answers.
In contrast, a larger $\alpha$ results in better performance in summarization task.
As the summarization task is more objective than the dialogue task, the results are more sensitive to noise, and hence we need a model that is more robust to the noise.

\section{Related Work}

\textbf{Preference Alignment of LLMs.}
The most representative paradigm of preference alignment is RLHF \cite{ziegler2019fine, ouyang2022training}, which involves training a reward model to capture human preferences and then steering LLMs towards producing high-reward responses through RL algorithms \cite{schulman2017proximal}.
However, in real applications, RL-based methods are complex and prone to instability during training \cite{dpo, rlhf-complex-1, rlhf-complex-2}.
Therefore, many recent studies have explored more straightforward and stable alternatives for RLHF \cite{rlhf-complex-2, dpo, song2023preference, wang2023aligning-v1, lin2023unlocking, li2023rain, wang2023aligning-v2, zhao2022calibrating}. 
Among these studies, the most promising direction is to use a contrastive or ranking loss to calibrate the likelihood of the output sequence.
Specifically, RRHF \cite{rlhf-complex-2} introduces a ranking loss to encourage larger likelihoods for better responses and smaller likelihoods for worse responses.
Sequence Likelihood Calibration (SLiC) \cite{zhao2022calibrating} uses a spectrum of calibration losses to align the model’s outputs with reference sequences in the latent space.
Besides, another important work is DPO \cite{dpo}, which implicitly optimizes the same objective as existing RLHF-based methods and enables human preference alignment directly with a simple cross-entropy loss.
In addition to the aforementioned methods using data in the form of $(\mathbf{x}, \mathbf{y}_1, \mathbf{y}_2, c)$, where $c$ is the preference label, some recent studies \cite{duan2024negating, ethayarajh2024kto, chen2024self} have also used data in the form of $(\mathbf{x}, \mathbf{y}, \mathbf{c})$, where $c$ is an annotation of the response $\mathbf{y}$, for preference alignment.

\textbf{Learning from Noisy Data.}
In the era of deep learning and LLMs, there is an urgent demand for large-scale training samples, and the cost of manually annotating or filtering data is often prohibitively expensive in most circumstances \cite{song2022learning}.
Therefore, techniques for learning from noise data have been increasingly important, which primarily fall into three categories.
The first category is sample-selection based methods \cite{ swayamdipta2020dataset, pleiss2020identifying, paul2021deep, sorscher2022beyond}, which identify high-quality samples before training and filter out the noisy or valueless samples.
For example, \cite{swayamdipta2020dataset} use the training dynamics to identify valuable samples, and \cite{liu2020early} indicates that the magnitude of the loss can be used to identify noisy samples from clean ones.
The second category is weighting-based methods, which assign greater weights for important samples and lesser weights for noisy samples within the loss function \cite{ren2018learning, han2022learning, shu2019meta}.
Besides, another important area of research is dedicated to the design of loss functions that are robust to noise \cite{ghosh2017robust, wang2019symmetric, zhang2018generalized}.
The findings in \cite{ghosh2017robust} indicate that the traditional cross-entropy loss is sensitive to the label noise, while symmetric loss functions are robust to such noise.
Furthermore, recent advances in LLMs have also underscored the essential role of data quality in both pre-training and supervised fine-tuning (SFT) phases of LLMs \cite{marion2023less, zhou2023lima, korbak2023pretraining}.

\section{Conclusion}

Robust preference optimization is a critical technique in the LLM alignment, as the noisy preferences are inevitable in practical scenarios.
Unlike existing methods, which rely on label smoothing or the use of external LLMs for sample selection, our approach introduces a robust preference alignment framework that unifies noise-tolerant model training and noisy sample filtering.
Specifically, a noise-tolerant regularization term is incorporated to prevent the model from over-fitting to noise.
Besides, we also demonstrate that the proposed regularization term plays a crucial role in distinguishing noisy samples from clean ones.
Furthermore, the rejection sampling technique is utilized to compensate for the potential information reduction caused by sample filtering stage, which brings additional performance improvement.
Extensive theoretical and empirical evidences are provided to demonstrate the effectiveness of our proposed ROPO framework.

\textbf{Limitations and broader impacts.}
For datasets with unknown noise proportion, the ROPO employs a pre-fixed sample filtering ratio, posing a potential risk of either over-filtering or insufficiently eliminating noisy data.
Therefore, one promising optimization direction of our approach is to filter out noisy preference samples with a data-adaptive ratio.
This work aims to improve the response quality of LLMs, which tends to benefit a wide range of LLM-based applications.

\bibliography{neurips_2024}
\bibliographystyle{plain}

\newpage
\appendix

\section{More Details about Experiments}\label{append:exp}

\subsection{Tasks and Datasets}\label{append:dataset}

We run experiments on two dialogue datasets (i.e., UltraFeedback Binarized and Alpaca Comparison) and one post summarization dataset (i.e., TL;DR).

\begin{itemize}[leftmargin=15pt]
    \item The UltraFeedback Binarized dataset\footnote{\url{https://huggingface.co/datasets/HuggingFaceH4/ultrafeedback_binarized}} is a pre-processed version of the UltraFeedback dataset \cite{cui2023ultrafeedback}, which contains 64,000 prompts and each prompt has four model responses from various LLMs.
    Based on the score assigned by GPT-4, \cite{tunstall2023zephyr} selects two responses for each prompt and construct UltraFeedback Binarized for the preference alignment.

    \item The Alpaca Comparison dataset contains 52,000 queries from the widely-used Stanford Alpaca dataset \cite{alpaca}.
    \cite{peng2023instruction} generates several responses using GPT-4 and other LLMs including text-davinci-003 to each query and employs GPT-4 to assign a score for each response.
    
    \item In the TL;DR dataset, each prompt is a forum from Reddit, and the model is required to summarize the given forum.
    Following \cite{dpo}, we use the Reddit TL;DR summarization dataset \cite{tldr-1} along with human preferences collected by \cite{tldr-2}.
\end{itemize}

\subsection{Baselines, Models, and Hyperparameters}\label{append:baselines}

\textbf{Baselines.}
Our baselines are DPO \cite{dpo}, IPO \cite{ipo}, and two approaches that use the label smoothing technique to alleviate the impact of noise, i.e., rDPO \cite{rdpo} and cDPO \cite{cdpo}.

Specifically, given a preference data $(\mathbf{x}, \mathbf{y}_1, \mathbf{y}_2)$ with the ranking label $\mathbf{y}_1 \succ \mathbf{y}_2 \mid \mathbf{x}$, the objectives of our baselines are
\begin{align}
    &\ell_{\rm dpo} = -\log \sigma\bigg(\beta \log \frac{\pi_{\theta}(\mathbf{y}_1 \mid \mathbf{x})}{\pi_{\rm ref}(\mathbf{y}_1 \mid \mathbf{x})} - \beta \log \frac{\pi_{\theta}(\mathbf{y}_2 \mid \mathbf{x})}{\pi_{\rm ref}(\mathbf{y}_2 \mid \mathbf{x})}\bigg),\\
    &\ell_{\rm ipo} = \left( \log \frac{\pi_{\theta}(\mathbf{y}_1 \mid \mathbf{x})}{\pi_{\rm ref}(\mathbf{y}_1 \mid \mathbf{x})} - \log \frac{\pi_{\theta}(\mathbf{y}_2 \mid \mathbf{x})}{\pi_{\rm ref}(\mathbf{y}_2 \mid \mathbf{x})}  - \frac{1}{2\beta} \right)^2,\\
    &\ell_{\rm rdpo} = -\frac{1-\varepsilon}{1 - 2\varepsilon} \log \sigma\bigg(\beta \log \frac{\pi_{\theta}(\mathbf{y}_1 \mid \mathbf{x})}{\pi_{\rm ref}(\mathbf{y}_1 \mid \mathbf{x})} - \beta \log \frac{\pi_{\theta}(\mathbf{y}_2 \mid \mathbf{x})}{\pi_{\rm ref}(\mathbf{y}_2 \mid \mathbf{x})}\bigg)\nonumber\\
    &\quad\quad\quad\,\,\,\,\,+\frac{\varepsilon}{1-2\varepsilon} \log \sigma\bigg(\beta \log \frac{\pi_{\theta}(\mathbf{y}_2 \mid \mathbf{x})}{\pi_{\rm ref}(\mathbf{y}_2 \mid \mathbf{x})} - \beta \log \frac{\pi_{\theta}(\mathbf{y}_1 \mid \mathbf{x})}{\pi_{\rm ref}(\mathbf{y}_1 \mid \mathbf{x})}\bigg),\\
    &\ell_{\rm cdpo} = -\varepsilon \log \sigma\bigg(\beta \log \frac{\pi_{\theta}(\mathbf{y}_1 \mid \mathbf{x})}{\pi_{\rm ref}(\mathbf{y}_1 \mid \mathbf{x})} - \beta \log \frac{\pi_{\theta}(\mathbf{y}_2 \mid \mathbf{x})}{\pi_{\rm ref}(\mathbf{y}_2 \mid \mathbf{x})}\bigg)\nonumber\\
    &\quad\quad\quad\,\,\,\,\,-(1-\varepsilon) \log \sigma\bigg(\beta \log \frac{\pi_{\theta}(\mathbf{y}_2 \mid \mathbf{x})}{\pi_{\rm ref}(\mathbf{y}_2 \mid \mathbf{x})} - \beta \log \frac{\pi_{\theta}(\mathbf{y}_1 \mid \mathbf{x})}{\pi_{\rm ref}(\mathbf{y}_1 \mid \mathbf{x})}\bigg),
\end{align}
where $\varepsilon\in(0,\frac{1}{2})$ and $\beta\in(0,1)$ are hyperparameters.

\textbf{Models.}
We use Mistral-7B \cite{jiang2023mistral} and Llama-2-7B \cite{llama2} as base models for all baselines and datasets.
On UFB, we use Zephyr-7B-SFT-$\beta$ \cite{tunstall2023zephyr} as the SFT model for experiments with Mistral-7B, and adopt the result of Zephyr-7B-$\beta$ \cite{tunstall2023zephyr} on AlpacaEval (90.60) as the performance of DPO under no artificial noise.
In other cases, we fine-tune base models on the preferred responses (SFT targets) to form the SFT models.

\textbf{Hyperparameters.}
We run all experiments on 16 NVIDIA A100 GPUs (80 GB).
Unless otherwise noted, we use a global batch size of 512 to train all models.
For all hyperparameters except for $\varepsilon$ of label smoothing, we search for the best one on each dataset without artificial noise and use the same setting for 20\% and 40\% artificial noise.

For all methods, we search the best learning rate in \{1e-5, 5e-6, 1e-6, 5e-7, 1e-7\} and the best $\beta$ in \{0.1, 0.5\}.
We find that the best performing learning rate is 1e-6, and the best $\beta$ for dialogue and post summarization are 0.1 and 0.5, respectively.
This conclusion is consistent with that in \cite{dpo}.

For ROPO, we search the best $\alpha$ in \{6, 14, 30\}, which makes $\frac{4\alpha}{(1+\alpha)^2}$ be around $\frac{1}{2}, \frac{1}{4}, \frac{1}{8}$, respectively, and search the best $\rho$ in \{0.1, 0.2, 0.3\}.
We find that $(\alpha, \rho)=(14, 0.2)$ performs the best when there is no artificial noise.
In Section \ref{sec:ablation} we provide the ablation studies on $\alpha$ and $\rho$.
We set $K=3$ for the rejection sampling.
For rDPO and cDPO, we search the best $\varepsilon$ in \{0,1, 0.2, 0.3, 0.4\} for each dataset and each proportion of artificial noise.

\subsection{Evaluation}\label{append:evaluation}

For models trained on UFB and Alpaca Comparison, we evaluate them on the AlpacaEval benchmark \cite{alpaca_eval}---a widely used dialogue benchmark---by comparing their outputs with those of text-davinci-003 (recommended by the benchmark for comparison).
AlpacaEval contains 805 queries in various domains and exhibit a strong concordance with ground truth human annotators.
For TL;DR, we randomly select 500 queries from the test split of it and evaluate ROPO and baselines by comparing their outputs with the chosen responses (SFT targets) for the queries.

Following existing studies \cite{dpo, tunstall2023zephyr}, we employ GPT-4 as the referee to conduct head-to-head comparisons, using the win rate as the metric.
On AlpacaEval, we conduct evaluations using the API provided by AlpacaEval.
On TL;DR, we use the following prompt, which is similar to that used by AlpacaEval, to conduct GPT-4 evaluation.

\begin{center}
\noindent
\fbox{
    \parbox{.99\linewidth}{

        \texttt{You are a helpful assistant that ranks models by the quality of their summaries of given forum posts.}\\\

        \texttt{I want you to create a leaderboard of different of large-language models. To do so, I will give you the instructions (forum posts) given to the models, and the responses of two models. Please rank the models based on which responses would be preferred by humans.}\\\
        
        \texttt{Here is the post:}
        
        \texttt{<Forum Post>}\\\
        
        \texttt{Here are the outputs of the models:}\\\
        \texttt{Model 1: <Summary 1>}\\\
        \texttt{Model 2: <Summary 2>}\\\
        
        \texttt{Now please rank the models by the quality of their answers, so that the model with rank 1 has the best output. Please provide the ranking that the majority of humans would give. Your response should use the format:}\\\
        \texttt{Better: <Model 1 or Model 2>}
}
}
\end{center}

\section{Mathematical Derivations and Theoretical Analysis}\label{append:math}

\subsection{Proof of Theorem \ref{thm:primal_problem}}

\begin{proof}
    As $\sum_{i=1}^N w_i = N_\rho$ is a hyperplane and $w_i \in [0,1]$ for $i=1,\ldots,N$, $S\triangleq\{\mathbf{w}: w_i \in [0,1], \sum_{i=1}^N w_i = N_\rho\}$ is compact.
    Because $\Theta$ is compact, $\Theta \times S$ is compact.
    Therefore, the continuous $\frac{1}{N} \sum_{i=1}^N w_i \ell \left(\theta; \mathbf{x}^{(i)}, \mathbf{y}_1^{(i)}, \mathbf{y}_2^{(i)}, \hat{c}^{(i)}, \pi_\theta\right)$ admits an optimal solution $(\theta^*, \mathbf{w}^*)$ on $\Theta \times S$.

    Assume that $\ell\left(\theta^*;\mathbf{x}^{(i_1)}, \mathbf{y}^{(i_1)}_1, \mathbf{y}^{(i_1)}_2, \pi_{\theta^*}\right) < \cdots < \ell\left(\theta^*;\mathbf{x}^{(i_N)}, \mathbf{y}^{(i_N)}_1, \mathbf{y}^{(i_N)}_2, \pi_{\theta^*}\right)$ but with $w^*_{i_j} < 1$ for some $1\leq j \leq N_\rho$.
    Then, we have
    \begin{align}
        \sum_{k=1}^{N_\rho} w^*_{i_k} < 1 + (N_\rho - 1) = N_\rho,
    \end{align}
    hence there exists $w_{i_l}^* > 0$ for some $N_\rho < l \leq N$.
    By letting $w'_{i_j} = 1$, $w'_{i_l} = w^*_{i_j} + w^*_{i_l} - 1$, and $w'_{i_k} = w^*_{i_k}$ for $k \neq j,l$, we have $\sum_{k=1}^N w'_{i_k} = 1$ and
    \begin{align}
        \frac{1}{N}\sum_{i=1}^N w_i' \ell \left(\theta^*; \mathbf{x}^{(i)}, \mathbf{y}_1^{(i)}, \mathbf{y}_2^{(i)}, \hat{c}^{(i)}, \pi_{\theta^*}\right) ={}& \frac{1}{N} \sum_{k\neq j, l} w_{i_k}' \ell \left(\theta^*; \mathbf{x}^{(i_k)}, \mathbf{y}_1^{(i_k)}, \mathbf{y}_2^{(i_k)}, \hat{c}^{(i_k)}, \pi_{\theta^*}\right)\nonumber\\
        &+  w'_{i_j} \ell \left(\theta^*; \mathbf{x}^{(i_j)}, \mathbf{y}_1^{(i_j)}, \mathbf{y}_2^{(i_j)}, \hat{c}^{(i_j)}, \pi_{\theta^*}\right)\nonumber\\
        &+ w'_{i_l} \ell \left(\theta^*; \mathbf{x}^{(i_l)}, \mathbf{y}_1^{(i_l)}, \mathbf{y}_2^{(i_l)}, \hat{c}^{(i_l)}, \pi_{\theta^*}\right) \nonumber\\
        <{}& \frac{1}{N} \sum_{k\neq j, l} w_{i_k}^* \ell \left(\theta^*; \mathbf{x}^{(i_k)}, \mathbf{y}_1^{(i_k)}, \mathbf{y}_2^{(i_k)}, \hat{c}^{(i_k)}, \pi_{\theta^*}\right)\nonumber\\
        &+  w^*_{i_j} \ell \left(\theta^*; \mathbf{x}^{(i_j)}, \mathbf{y}_1^{(i_j)}, \mathbf{y}_2^{(i_j)}, \hat{c}^{(i_j)}, \pi_{\theta^*}\right)\nonumber\\
        &+ w^*_{i_l} \ell \left(\theta^*; \mathbf{x}^{(i_l)}, \mathbf{y}_1^{(i_l)}, \mathbf{y}_2^{(i_l)}, \hat{c}^{(i_l)}, \pi_{\theta^*}\right) \nonumber\\
        ={}& \frac{1}{N}\sum_{i=1}^N w_i^* \ell \left(\theta^*; \mathbf{x}^{(i)}, \mathbf{y}_1^{(i)}, \mathbf{y}_2^{(i)}, \hat{c}^{(i)}, \pi_{\theta^*}\right),
    \end{align}
    which leads to a contradiction.
    Therefore, we must have $w^*_{i_k} = 1$ for $1\le k \le N_\rho$ and $w^*_{i_k} = 0$ for $N_\rho < k \le N$.
\end{proof}

\subsection{Proof of Theorem \ref{thm:dpo_noise_tolerance}}

\begin{proof}
    For $\ell = \ell_{\rm dpo}$, we have
    \begin{align}\label{eqn:exp_dpo}
        &\mathbb{E}_{(\mathbf{x}, \mathbf{y}_1, \mathbf{y}_2, \hat{c}) \sim \mathcal{D}_\eta} [\ell(\theta; \mathbf{x}, \mathbf{y}_1, \mathbf{y}_2, \hat{c}, \pi_\theta)]\nonumber\\
        ={}& \mathbb{E}_{(\mathbf{x}, \mathbf{y}_1, \mathbf{y}_2)} \mathbb{E}_{c \mid \mathbf{x}, \mathbf{y}_1, \mathbf{y}_2} \mathbb{E}_{\hat{c} \mid \mathbf{x}, \mathbf{y}_1, \mathbf{y}_2, c} [\ell(\theta; \mathbf{x}, \mathbf{y}_1, \mathbf{y}_2, \hat{c}, \pi_\theta)] \nonumber\\
        ={}& \mathbb{E}_{(\mathbf{x}, \mathbf{y}_1, \mathbf{y}_2)} \bigg[\left(P^*(\mathbf{y}_1 \succ \mathbf{y}_2 \mid \mathbf{x}) (1-\eta) + (1-P^*(\mathbf{y}_1 \succ \mathbf{y}_2 \mid \mathbf{x}))\eta \right)\cdot \ell(\theta; \mathbf{x}, \mathbf{y}_1, \mathbf{y}_2, 0, \pi_\theta)\nonumber\\
        &\quad\quad\quad\quad + \left(P^*(\mathbf{y}_1 \succ \mathbf{y}_2 \mid \mathbf{x}) \eta + (1-P^*(\mathbf{y}_1 \succ \mathbf{y}_2 \mid \mathbf{x}))(1-\eta)\right)\cdot \ell(\theta; \mathbf{x}, \mathbf{y}_1, \mathbf{y}_2, 1, \pi_\theta) \bigg] \nonumber\\
        ={}& \mathbb{E}_{(\mathbf{x}, \mathbf{y}_1, \mathbf{y}_2)} \bigg[ -\left( P^*(\mathbf{y}_1 \succ \mathbf{y}_2 \mid \mathbf{x}) + \eta - 2 P^*(\mathbf{y}_1 \succ \mathbf{y}_2 \mid \mathbf{x}) \eta   \right) \log P_\theta(\mathbf{y}_1 \succ \mathbf{y}_2 \mid \mathbf{x}) \nonumber\\
        &\quad\quad\quad\quad -\left( 2P^*(\mathbf{y}_1 \succ \mathbf{y}_2 \mid \mathbf{x})\eta + 1 - P^*(\mathbf{y}_1 \succ \mathbf{y}_2 \mid \mathbf{x}) - \eta  \right) \log (1 - P_\theta(\mathbf{y}_1 \succ \mathbf{y}_2 \mid \mathbf{x})) \bigg].
    \end{align}
    Consider
    \begin{align}
        f(p) = -(p^* + \eta - 2p^* \eta) \log p - (2p^*\eta + 1 - p^* - \eta) \log (1-p),
    \end{align}
    we have
    \begin{align}
        f'(p) = -\frac{p^* + \eta - 2 p^*\eta}{p} + \frac{2p^* \eta + 1 - p^* - \eta}{1 - p}.
    \end{align}
    From $f'(p)$ we know that $f$ decrease when $p\leq p^* + \eta - 2p^* \eta$ and increases when $p\ge p^* + \eta - 2p^* \eta$, which means that $f$ reaches its minimum at $p_0 = p^* + (1 - 2p^*)\eta$.

    Therefore, Eq. (\ref{eqn:exp_dpo}) reaches its minimum when
    \begin{align}\label{eqn:dpo_p}
        P_{\theta^*_\eta}(\mathbf{y}_1 \succ \mathbf{y}_2 \mid \mathbf{x}) = P^*(\mathbf{y}_1 \succ \mathbf{y}_2 \mid \mathbf{x}) + (1 - 2P^*(\mathbf{y}_1 \succ \mathbf{y}_2 \mid \mathbf{x}))\eta
    \end{align}
    for any $(\mathbf{x}, \mathbf{y}_1, \mathbf{y}_2)$.
    Specifically, for $\eta=0$, we have $P_{\theta^*}(\mathbf{y}_1 \succ \mathbf{y}_2 \mid \mathbf{x}) = P^*(\mathbf{y}_1 \succ \mathbf{y}_2 \mid \mathbf{x})$, which leas to
    \begin{align}
        \big|P_{\theta^*_\eta}(\mathbf{y}_1 \succ \mathbf{y}_2 \mid \mathbf{x}) - P_{\theta^*}(\mathbf{y}_1 \succ \mathbf{y}_2 \mid \mathbf{x})\big| = 2 \eta \big| P^*(\mathbf{y}_1 \succ \mathbf{y}_2 \mid \mathbf{x}) - 1/2\big|.
    \end{align}
\end{proof}

\subsection{Proof of Theorem \ref{thm:dpo_filter}}
\begin{proof}
    For samples $(\mathbf{x}^{(1)}, \mathbf{y}^{(1)}_1, \mathbf{y}^{(1)}_2, \hat{c}^{(1)} = c^{(1)})$ and $(\mathbf{x}^{(2)}, \mathbf{y}^{(2)}_1, \mathbf{y}^{(2)}_2, \hat{c}^{(2)} = 1 - c^{(2)})$, according to Eq. (\ref{eqn:dpo_p}), we have
    \begin{align}
        P_{\theta^*_\eta} \left(  \mathbf{x}^{(1)}, \mathbf{y}_1^{(1)}, \mathbf{y}_2^{(1)}, \hat{c}^{(1)} \right) ={}& P_{\theta^*_\eta} \left(  \mathbf{x}^{(1)}, \mathbf{y}_1^{(1)}, \mathbf{y}_2^{(1)}, c^{(1)} \right)\nonumber\\
        ={}& P^*( c^{(1)} ) + (1 - 2P^*(c^{(1)}))\eta
    \end{align}
    and
    \begin{align}
        P_{\theta^*_\eta} \left(  \mathbf{x}^{(2)}, \mathbf{y}_1^{(2)}, \mathbf{y}_2^{(2)}, \hat{c}^{(2)} \right) ={}& P_{\theta^*_\eta} \left(  \mathbf{x}^{(2)}, \mathbf{y}_1^{(2)}, \mathbf{y}_2^{(2)}, 1 - c^{(2)} \right)\\
        ={}& P^*( 1 - c^{(2)} ) + (1 - 2P^*(1 - c^{(2)}))\eta \nonumber\\
        ={}& 1 - P^*(c^{(2)}) + (2P^*(c^{(2)}) - 1)\eta.
    \end{align}
    Therefore, to ensure that
    \begin{align}
        \ell_{\rm dpo}\left(\mathbf{x}^{(1)}, \mathbf{y}^{(1)}_1, \mathbf{y}^{(1)}_2, \hat{c}^{(1)}\right) - \ell_{\rm dpo}\left(\mathbf{x}^{(2)}, \mathbf{y}^{(2)}_1, \mathbf{y}^{(2)}_2, \hat{c}^{(2)}\right) < 0,
    \end{align}
    we must have
    \begin{align}
        -\log\left( P^*( c^{(1)} ) + (1 - 2P^*(c^{(1)}))\eta - \varepsilon \right) < -\log \left(  1 - P^*(c^{(2)}) + (2P^*(c^{(2)}) - 1)\eta + \varepsilon \right),
    \end{align}
    which is equivalent to
    \begin{align}
        \varepsilon < \frac{1 - 2\eta}{2} \left( P^*( c^{(1)} ) + P^*( c^{(2)} ) -1 \right).
    \end{align}
\end{proof}

\subsection{Detailed Derivation of Eq. (\ref{eqn:decomposition})}\label{append:decomposition}

From the definition of $w_{\rm ropo}$ we have
\begin{align}
    w_{\rm ropo} = \frac{4\alpha}{(1+\alpha)^2} \sigma(\Delta(\mathbf{y}_2, \mathbf{y}_1, \mathbf{x})) + \frac{4\alpha^2}{(1+\alpha)^2} \sigma(\Delta(\mathbf{y}_2, \mathbf{y}_1, \mathbf{x}))\sigma(\Delta(\mathbf{y}_1, \mathbf{y}_2, \mathbf{x})).
\end{align}
According to Eq. (\ref{eqn:dpo_gradient}) we know that
\begin{align}
    -\int \beta \frac{4\alpha}{(1+\alpha)^2} \sigma(\Delta(\mathbf{y}_2, \mathbf{y}_1, \mathbf{x}))\nabla \log\frac{\pi_\theta(\mathbf{y}_1 \mid \mathbf{x})}{\pi_\theta(\mathbf{y}_2 \mid \mathbf{x})} \,{\rm d}\theta = \frac{4\alpha}{(1+\alpha)^2} \ell_{\rm dpo}.
\end{align}
Beside, note that for $\sigma(x) = \frac{e^x}{1 + e^x}$, we have
\begin{align}
    \sigma'(x) = \left( \frac{e^{x}}{1+e^{x}} \right)' = \frac{e^x(1 + e^x) - e^x \cdot e^x}{(1+e^x)^2} = \frac{e^x}{(1+e^x)^2} = \frac{e^x}{1+ e^x}\cdot\frac{1}{1+e^x} = \sigma(x)\sigma(-x)
\end{align}
and
\begin{align}
    \sigma'(-x) = -\sigma(x) \sigma(-x).
\end{align}
Letting
\begin{align}
    t(\theta) = \beta \log \frac{\pi_\theta (\mathbf{y}_1 \mid \mathbf{x})}{\pi_{\rm ref}(\mathbf{y}_1 \mid \mathbf{x})} - \beta \log \frac{\pi_\theta (\mathbf{y}_2 \mid \mathbf{x})}{\pi_{\rm ref}(\mathbf{y}_2 \mid \mathbf{x})},
\end{align}
we have
\begin{align}
    \nabla_\theta t(\theta) = \beta \nabla \log \frac{\pi_\theta(\mathbf{y}_1\mid \mathbf{x})}{\pi_\theta(\mathbf{y}_2 \mid \mathbf{x})}
\end{align}
Hence,
\begin{align}
    &-\frac{4\alpha^2}{(1+\alpha)^2} \int \beta \sigma(\Delta(\mathbf{y}_2, \mathbf{y}_1, \mathbf{x}))\sigma(\Delta(\mathbf{y}_1, \mathbf{y}_2, \mathbf{x})) \nabla \log\frac{\pi_\theta(\mathbf{y}_1 \mid \mathbf{x})}{\pi_\theta(\mathbf{y}_2 \mid \mathbf{x})}\,{\rm d}\theta\nonumber \\
    ={}& \frac{4\alpha^2}{(1+\alpha)^2}\int \bigg(-\sigma(t(\theta)) \sigma(-t(\theta))\bigg) \cdot \bigg( \beta \nabla \log \frac{\pi_\theta(\mathbf{y}_1\mid \mathbf{x})}{\pi_\theta(\mathbf{y}_2 \mid \mathbf{x})}\bigg) \,{\rm d}\theta\nonumber\\
    ={}& \frac{4\alpha^2}{(1+\alpha)^2}\int \nabla_{t(\theta)} \sigma(-t(\theta)) \cdot \nabla_\theta t(\theta)\,{\rm d}\theta \nonumber\\
    ={}& \frac{4\alpha^2}{(1+\alpha)^2}\int \nabla_\theta \sigma(-t(\theta)) \,{\rm d}\theta\nonumber\\
    ={}& \frac{4\alpha^2}{(1+\alpha)^2}\sigma(-t(\theta))\nonumber\\
    ={}& \frac{4\alpha^2}{(1+\alpha)^2} \cdot \sigma\left(\beta \log \frac{\pi_\theta (\mathbf{y}_2 \mid \mathbf{x})}{\pi_{\rm ref}(\mathbf{y}_2 \mid \mathbf{x})} - \beta \log \frac{\pi_\theta (\mathbf{y}_1 \mid \mathbf{x})}{\pi_{\rm ref}(\mathbf{y}_1 \mid \mathbf{x})}\right),
\end{align}
where we omit the constant term of the primitive function.

\subsection{Proof of Theorem \ref{thm:reg-robust}}

\begin{proof}
    For $\ell = \ell_{\rm reg}$, we have
    \begin{align}\label{eqn:exp_reg}
        &\mathbb{E}_{(\mathbf{x}, \mathbf{y}_1, \mathbf{y}_2, \hat{c}) \sim \mathcal{D}_\eta} [\ell(\theta; \mathbf{x}, \mathbf{y}_1, \mathbf{y}_2, \hat{c}, \pi_\theta)]\nonumber\\
        ={}& \mathbb{E}_{(\mathbf{x}, \mathbf{y}_1, \mathbf{y}_2)} \mathbb{E}_{c \mid \mathbf{x}, \mathbf{y}_1, \mathbf{y}_2} \mathbb{E}_{\hat{c} \mid \mathbf{x}, \mathbf{y}_1, \mathbf{y}_2, c} [\ell(\theta; \mathbf{x}, \mathbf{y}_1, \mathbf{y}_2, \hat{c}, \pi_\theta)] \nonumber\\
        ={}& \mathbb{E}_{(\mathbf{x}, \mathbf{y}_1, \mathbf{y}_2)} \bigg[\left(P^*(\mathbf{y}_1 \succ \mathbf{y}_2 \mid \mathbf{x}) (1-\eta) + (1-P^*(\mathbf{y}_1 \succ \mathbf{y}_2 \mid \mathbf{x}))\eta \right)\cdot \ell(\theta; \mathbf{x}, \mathbf{y}_1, \mathbf{y}_2, 0, \pi_\theta)\nonumber\\
        &\quad\quad\quad\quad + \left(P^*(\mathbf{y}_1 \succ \mathbf{y}_2 \mid \mathbf{x}) \eta + (1-P^*(\mathbf{y}_1 \succ \mathbf{y}_2 \mid \mathbf{x}))(1-\eta)\right)\cdot \ell(\theta; \mathbf{x}, \mathbf{y}_1, \mathbf{y}_2, 1, \pi_\theta) \bigg] \nonumber\\
        ={}& \mathbb{E}_{(\mathbf{x}, \mathbf{y}_1, \mathbf{y}_2)} \bigg[ \left( P^*(\mathbf{y}_1 \succ \mathbf{y}_2 \mid \mathbf{x}) + \eta - 2 P^*(\mathbf{y}_1 \succ \mathbf{y}_2 \mid \mathbf{x}) \eta   \right) (1 -P_\theta(\mathbf{y}_1 \succ \mathbf{y}_2 \mid \mathbf{x})) \nonumber\\
        &\quad\quad\quad\quad +\left( 2P^*(\mathbf{y}_1 \succ \mathbf{y}_2 \mid \mathbf{x})\eta + 1 - P^*(\mathbf{y}_1 \succ \mathbf{y}_2 \mid \mathbf{x}) - \eta  \right) P_\theta(\mathbf{y}_1 \succ \mathbf{y}_2 \mid \mathbf{x}) \bigg].
    \end{align}
    Consider
    \begin{align}
        f(p) ={}& (p^* + \eta - 2p^* \eta) (1-p) + (2p^*\eta + 1 - p^* - \eta) p\nonumber\\
        ={}& (1-2\eta) (1-2p^*) p + (p^* + \eta - 2p^*\eta).
    \end{align}
    Therefore, when $p^* > 1/2$, $f(p)$ reaches its minimum at $p=1$; when $p^* < 1/2$, $f(p)$ reaches its minimum at $p=0$.
    This means that the optimal point of $f(p)$ is $p_0 = \mathbb{I}(p^* > 1/2)$.

    Therefore, Eq. (\ref{eqn:exp_reg}) reaches its minimum when
    \begin{align}\label{eqn:reg_p}
        P_{\theta^*_\eta}(\mathbf{y}_1 \succ \mathbf{y}_2 \mid \mathbf{x}) = \mathbb{I}\left( P^*(\mathbf{y}_1 \succ \mathbf{y}_2 \mid \mathbf{x}) > \frac{1}{2} \right)
    \end{align}
    for any $(\mathbf{x}, \mathbf{y}_1, \mathbf{y}_2)$.
    Obviously, we have
    \begin{align}
        P_{\theta^*_\eta}(\mathbf{y}_1 \succ \mathbf{y}_2 \mid \mathbf{x}) = P_{\theta^*}(\mathbf{y}_1 \succ \mathbf{y}_2 \mid \mathbf{x}).
    \end{align}
\end{proof}

\subsection{Proof of Theorem \ref{thm:reg-dist}}

\begin{proof}
    For samples $(\mathbf{x}^{(1)}, \mathbf{y}^{(1)}_1, \mathbf{y}^{(1)}_2, \hat{c}^{(1)} = c^{(1)})$ and $(\mathbf{x}^{(2)}, \mathbf{y}^{(2)}_1, \mathbf{y}^{(2)}_2, \hat{c}^{(2)} = 1 - c^{(2)})$.
    Without loss of generality, we only need to consider two cases: (1) $c^{(1)} = c^{(2)} = 0$ and (2) $c^{(1)} = 0, c^{(2)} = 1$.
    For the first case, we have
    \begin{align}
        \ell_{\rm reg} \left(\mathbf{x}^{(1)}, \mathbf{y}^{(1)}_1, \mathbf{y}^{(1)}_2, \hat{c}^{(1)}\right) = P_\theta (\mathbf{y}_2^{(1)} \succ \mathbf{y}_1^{(1)} \mid \mathbf{x}) \in [0, \varepsilon)
    \end{align}
    and
    \begin{align}
        \ell_{\rm reg} \left(\mathbf{x}^{(2)}, \mathbf{y}^{(2)}_1, \mathbf{y}^{(2)}_2, \hat{c}^{(2)}\right) = P_\theta (\mathbf{y}_1^{(2)} \succ \mathbf{y}_2^{(2)} \mid \mathbf{x}) \in (1- \varepsilon, 1].
    \end{align}
    For the second case, we have
    \begin{align}
        \ell_{\rm reg} \left(\mathbf{x}^{(1)}, \mathbf{y}^{(1)}_1, \mathbf{y}^{(1)}_2, \hat{c}^{(1)}\right) = P_\theta (\mathbf{y}_2^{(1)} \succ \mathbf{y}_1^{(1)} \mid \mathbf{x}) \in [0, \varepsilon)
    \end{align}
    and 
    \begin{align}
        \ell_{\rm reg} \left(\mathbf{x}^{(2)}, \mathbf{y}^{(2)}_1, \mathbf{y}^{(2)}_2, \hat{c}^{(2)}\right) = P_\theta (\mathbf{y}_2^{(2)} \succ \mathbf{y}_1^{(2)} \mid \mathbf{x}) \in (1- \varepsilon, 1].
    \end{align}
    Therefore, to ensure that
    \begin{align}
        \ell_{\rm reg} \left(\mathbf{x}^{(1)}, \mathbf{y}^{(1)}_1, \mathbf{y}^{(1)}_2, \hat{c}^{(1)}\right) < \ell_{\rm reg} \left(\mathbf{x}^{(2)}, \mathbf{y}^{(2)}_1, \mathbf{y}^{(2)}_2, \hat{c}^{(2)}\right),
    \end{align}
    we must have $\varepsilon < \frac{1}{2}$.
\end{proof}

\subsection{rDPO and cDPO Are Not Noise-Tolerant In Most Cases}\label{append:rdpo-tolerant}

\begin{proof}

    According to Lemma 3.2 in \cite{rdpo}, the noise-tolerance of rDPO is only guaranteed when the proportion of noise, i.e., $\eta_0$, exactly equals the hyperparameter $\varepsilon$.

    Next we show that $\ell_{\rm cdpo}$ is not noise-tolerant for $\varepsilon \in (0, \frac{1}{2})$.
    Let
    \begin{align*}
        &\mathcal{L}_{\rm cdpo}(\theta) = \mathbb{E}_{(\mathbf{x}, \mathbf{y}_1, \mathbf{y}_2, c) \sim \mathcal{D}}[\ell_{\rm cdpo}(\theta; \mathbf{x}, \mathbf{y}_1, \mathbf{y}_2, c, \pi_\theta)],\\
        &\mathcal{L}^{\eta_0}_{\rm cdpo}(\theta) = \mathbb{E}_{(\mathbf{x}, \mathbf{y}_1, \mathbf{y}_2, \hat{c}) \sim \mathcal{D}_{\eta_0}}[\ell_{\rm cdpo}(\theta; \mathbf{x}, \mathbf{y}_1, \mathbf{y}_2, \hat{c}, \pi_\theta)],
    \end{align*}
    and assume that $\theta^*$ and $\theta_{\eta_0}^*$ are the minimizers of $\mathcal{L}_{\rm cdpo}$ and $\mathcal{L}_{\rm cdpo}^{\eta_0}$, respectively.
    For any $\theta$ in the space of parameters, we have
    \begin{align}\label{eqn:c-dpo_risk}
        &\mathcal{L}^{\eta_0}_{\rm cdpo}(\theta) \nonumber\\
        ={}& \mathbb{E}_{(\mathbf{x}, \mathbf{y}_1, \mathbf{y}_2,c)\sim\mathcal{D}}\mathbb{E}_{\hat{c}\mid (\mathbf{x}, \mathbf{y}_1, \mathbf{y}_2, c)}[\ell_{\rm cdpo}(\theta;\mathbf{x}, \mathbf{y}_1, \mathbf{y}_2,\hat{c},\pi_\theta)]\nonumber \\
        ={}& \mathbb{E}_{(\mathbf{x}, \mathbf{y}_1, \mathbf{y}_2,c)\sim\mathcal{D}}[(1-{\eta_0}) \ell_{\rm cdpo}(\theta;\mathbf{x}, \mathbf{y}_1, \mathbf{y}_2,c,\pi_\theta)\nonumber + {\eta_0} \ell_{\rm cdpo}(\theta;\mathbf{x}, \mathbf{y}_1, \mathbf{y}_2,1-c,\pi_\theta)]\nonumber\\
        ={}& (1-\eta_0) \mathcal{L}_{\rm cdpo}(\theta) + \eta_0 \mathbb{E}_{(\mathbf{x}, \mathbf{y}_1, \mathbf{y}_2,c)\sim\mathcal{D}}[\ell_{\rm cdpo}(\theta;\mathbf{x}, \mathbf{y}_1, \mathbf{y}_2,1-c,\pi_\theta)].
    \end{align}
    Next, we give a counter-example to show that $\ell_{\rm cdpo}$ is not noise-tolerant.
    Suppose that
    \begin{align}
        P\left((\mathbf{x}, \mathbf{y}_1, \mathbf{y}_2) = (\mathbf{x}^{(0)}, \mathbf{y}_1^{(0)}, \mathbf{y}_2^{(0)})\right) = 1\quad{\rm and}\quad \mathbf{y}_1^{(0)} \succ \mathbf{y}_2^{(0)} \mid \mathbf{x}^{(0)},
    \end{align}
    where $\mathbf{x}^{(0)}$ is a fixed input and $(\mathbf{y}_1^{(0)}, \mathbf{y}_2^{(0)})$ is a fixed pair of responses.
    Hence Eq. (\ref{eqn:c-dpo_risk}) becomes
    \begin{align}\label{eqn:counter-example-c-dpo}
        &\mathcal{L}^{\eta_0}_{\rm cdpo}(\theta) \nonumber\\
        ={}& (2\varepsilon\eta_0 -\eta_0 - \varepsilon)\log \sigma\left( \beta \log \frac{\pi_\theta(\mathbf{y}_1^{(0)} \mid \mathbf{x}^{(0)})}{\pi_{\rm ref}(\mathbf{y}_1^{(0)}\mid \mathbf{x}^{(0)})} -  \beta \log \frac{\pi_\theta(\mathbf{y}_2^{(0)}\mid \mathbf{x}^{(0)})}{\pi_{\rm ref}(\mathbf{y}_2^{(0)}\mid \mathbf{x}^{(0)})}\right)\nonumber\\
        {}&+ (\eta_0 + \varepsilon - 2\varepsilon\eta_0 - 1) \log \sigma\left( \beta \log \frac{\pi_\theta(\mathbf{y}_2^{(0)}\mid \mathbf{x}^{(0)})}{\pi_{\rm ref}(\mathbf{y}_2^{(0)}\mid \mathbf{x}^{(0)})} -  \beta \log \frac{\pi_\theta(\mathbf{y}_1^{(0)}\mid \mathbf{x}^{(0)})}{\pi_{\rm ref}(\mathbf{y}_1^{(0)}\mid \mathbf{x}^{(0)})}\right).
    \end{align}
    Let
    \begin{align}
        \Delta(\theta) = \beta \log \frac{\pi_\theta(\mathbf{y}_1^{(0)} \mid \mathbf{x}^{(0)})}{\pi_{\rm ref}(\mathbf{y}_1^{(0)}\mid \mathbf{x}^{(0)})} -  \beta \log \frac{\pi_\theta(\mathbf{y}_2^{(0)}\mid \mathbf{x}^{(0)})}{\pi_{\rm ref}(\mathbf{y}_2^{(0)}\mid \mathbf{x}^{(0)})},
    \end{align}
    then Eq. (\ref{eqn:counter-example-c-dpo}) becomes
    \begin{align}
        \mathcal{L}^{\eta_0}_{\rm cdpo}(\theta) = (2\varepsilon\eta_0 -\eta_0 - \varepsilon) \log\sigma(\Delta(\theta)) + (\eta_0 + \varepsilon - 2\varepsilon\eta_0 - 1) \log\sigma(-\Delta(\theta)).
    \end{align}
    We have
    \begin{align}
        \theta^* ={}& \mathop{\arg\min}\limits_{\theta\in\Theta} \,\,\mathcal{L}_{\rm cdpo} \nonumber\\
        ={}& \mathop{\arg\min}\limits_{\theta\in\Theta}\,\, -\varepsilon \log\sigma(\Delta(\theta)) - (1-\varepsilon) \log\sigma(\Delta(-\theta))\nonumber\\
        \in{}& \left\{\theta \in \Theta: \Delta(\theta) = \log\frac{\varepsilon}{1-\varepsilon}\right\},
    \end{align}
    and
    \begin{align}
        \theta^*_{\eta_0} ={}& \mathop{\arg\min}\limits_{\theta\in\Theta} \,\,\mathcal{L}_{\rm cdpo}^{\eta_0} \nonumber\\
        ={}& \mathop{\arg\min}\limits_{\theta\in\Theta}\,\, (2\varepsilon\eta_0 -\eta_0 - \varepsilon) \log\sigma(\Delta(\theta)) + (\eta_0 + \varepsilon - 2\varepsilon\eta_0 - 1) \log\sigma(-\Delta(\theta)) \nonumber\\
        \in{}& \left\{\theta \in \Theta: \Delta(\theta) = \log\frac{\eta_0+\varepsilon-2\varepsilon\eta_0}{1-\eta_0-\varepsilon+2\varepsilon\eta_0}\right\}.
    \end{align}
    Hence $\theta^* = \theta^*_{\eta_0}$ if and only if
    \begin{align}
        \frac{\varepsilon}{1-\varepsilon} = \frac{\eta_0+\varepsilon-2\varepsilon\eta_0}{1-\eta_0-\varepsilon+2\varepsilon\eta_0},
    \end{align}
    which means that $\varepsilon=\frac{1}{2}$.
    However, $\varepsilon\in(0, \frac{1}{2})$.
    Therefore, $\theta^* \neq \theta^*_{\eta_0}$ and thus $\ell_{\rm cdpo}$ is not noise-tolerant.
\end{proof}

\subsection{IPO Is Not Noise-Tolerant}\label{append:ipo-tolerant}
\begin{proof}
    Let
    \begin{align*}
        &\mathcal{L}_{\rm ipo}(\theta) = \mathbb{E}_{(\mathbf{x}, \mathbf{y}_1, \mathbf{y}_2, c) \sim \mathcal{D}}[\ell_{\rm ipo}(\theta; \mathbf{x}, \mathbf{y}_1, \mathbf{y}_2, c, \pi_\theta)],\\
        &\mathcal{L}^{\eta_0}_{\rm ipo}(\theta) = \mathbb{E}_{(\mathbf{x}, \mathbf{y}_1, \mathbf{y}_2, \hat{c}) \sim \mathcal{D}_{\eta_0}}[\ell_{\rm ipo}(\theta; \mathbf{x}, \mathbf{y}_1, \mathbf{y}_2, \hat{c}, \pi_\theta)],
    \end{align*}
    and assume that $\theta^*$ and $\theta_{\eta_0}^*$ are the minimizers of $\mathcal{L}_{\rm ipo}$ and $\mathcal{L}_{\rm ipo}^{\eta_0}$, respectively.
    For any $\theta$ in the space of parameters, we have
    \begin{align}\label{eqn:ipo_risk}
        &\mathcal{L}^{\eta_0}_{\rm ipo}(\theta) \nonumber\\
        ={}& \mathbb{E}_{(\mathbf{x}, \mathbf{y}_1, \mathbf{y}_2,c)\sim\mathcal{D}}\mathbb{E}_{\hat{c}\mid (\mathbf{x}, \mathbf{y}_1, \mathbf{y}_2, c)}[\ell_{\rm ipo}(\theta;\mathbf{x}, \mathbf{y}_1, \mathbf{y}_2,\hat{c},\pi_\theta)]\nonumber \\
        ={}& \mathbb{E}_{(\mathbf{x}, \mathbf{y}_1, \mathbf{y}_2,c)\sim\mathcal{D}}[(1-{\eta_0}) \ell_{\rm ipo}(\theta;\mathbf{x}, \mathbf{y}_1, \mathbf{y}_2,c,\pi_\theta)\nonumber + {\eta_0} \ell_{\rm ipo}(\theta;\mathbf{x}, \mathbf{y}_1, \mathbf{y}_2,1-c,\pi_\theta)]\nonumber\\
        ={}& (1-\eta_0) \mathcal{L}_{\rm ipo}(\theta) + \eta_0 \mathbb{E}_{(\mathbf{x}, \mathbf{y}_1, \mathbf{y}_2,c)\sim\mathcal{D}}[\ell_{\rm ipo}(\theta;\mathbf{x}, \mathbf{y}_1, \mathbf{y}_2,1-c,\pi_\theta)].
    \end{align}
    Next, we give a counter-example to show that $\ell_{\rm ipo}$ is not noise-tolerant.
    Suppose that
    \begin{align}
        P\left((\mathbf{x}, \mathbf{y}_1, \mathbf{y}_2) = (\mathbf{x}^{(0)}, \mathbf{y}_1^{(0)}, \mathbf{y}_2^{(0)})\right) = 1\quad{\rm and}\quad \mathbf{y}_1^{(0)} \succ \mathbf{y}_2^{(0)} \mid \mathbf{x}^{(0)},
    \end{align}
    where $\mathbf{x}^{(0)}$ is a fixed input and $(\mathbf{y}_1^{(0)}, \mathbf{y}_2^{(0)})$ is a fixed pair of responses.
    Hence Eq. (\ref{eqn:ipo_risk}) becomes
    \begin{align}\label{eqn:counter-example-ipo}
        &\mathcal{L}^{\eta_0}_{\rm ipo}(\theta) \nonumber\\
        ={}& (1-\eta_0) \left( \log \frac{\pi_\theta(\mathbf{y}_1^{(0)} \mid \mathbf{x}^{(0)})}{\pi_{\rm ref}(\mathbf{y}_1^{(0)}\mid \mathbf{x}^{(0)})} -  \log \frac{\pi_\theta(\mathbf{y}_2^{(0)}\mid \mathbf{x}^{(0)})}{\pi_{\rm ref}(\mathbf{y}_2^{(0)}\mid \mathbf{x}^{(0)})} - \frac{1}{2\beta}\right)^2\nonumber\\
        {}&+ \eta_0 \left( \log \frac{\pi_\theta(\mathbf{y}_2^{(0)}\mid \mathbf{x}^{(0)})}{\pi_{\rm ref}(\mathbf{y}_2^{(0)}\mid \mathbf{x}^{(0)})} -  \log \frac{\pi_\theta(\mathbf{y}_1^{(0)}\mid \mathbf{x}^{(0)})}{\pi_{\rm ref}(\mathbf{y}_1^{(0)}\mid \mathbf{x}^{(0)})}- \frac{1}{2\beta}\right)^2.
    \end{align}
    Let
    \begin{align}
        \Delta(\theta) = \log \frac{\pi_\theta(\mathbf{y}_1^{(0)} \mid \mathbf{x}^{(0)})}{\pi_{\rm ref}(\mathbf{y}_1^{(0)}\mid \mathbf{x}^{(0)})} -  \log \frac{\pi_\theta(\mathbf{y}_2^{(0)}\mid \mathbf{x}^{(0)})}{\pi_{\rm ref}(\mathbf{y}_2^{(0)}\mid \mathbf{x}^{(0)})},
    \end{align}
    then Eq. (\ref{eqn:counter-example-ipo}) becomes
    \begin{align}
        \mathcal{L}^{\eta_0}_{\rm ipo}(\theta) ={}& (1 - \eta_0) \left(\Delta(\theta)-\frac{1}{2\beta}\right)^2 + \eta_0 \left(-\Delta(\theta)-\frac{1}{2\beta}\right)^2\nonumber\\
        ={}& (\Delta(\theta))^2 + \frac{2\eta_0 - 1}{\beta} \Delta(\theta) + \frac{1}{4\beta^2},
    \end{align}
    which is a quadratic function.
    Hence
    \begin{align}
        \theta^*_{\eta_0} \in \left\{ \theta \in \Theta: \Delta(\theta) = \frac{1}{2\beta} - \frac{\eta_0}{\beta}\right\}.
    \end{align}
    However,
    \begin{align}
        \theta^* ={}& \mathop{\arg\min}\limits_{\theta\in\Theta} \,\,\mathcal{L}_{\rm ipo}\nonumber\\
        ={}& \mathop{\arg\min}\limits_{\theta\in\Theta} \left(\Delta(\theta)-\frac{1}{2\beta}\right)^2\nonumber\\
        \in{}& \left\{ \theta \in \Theta: \Delta(\theta) = \frac{1}{2\beta} \right\},
    \end{align}
    which means that $\theta^* \neq \theta^*_{\eta_0}$.
    Therefore, $\ell_{\rm ipo}$ is not noise-tolerant.
\end{proof}



\end{document}